%% file: acl_latex.tex
\documentclass[11pt]{article}

\usepackage[preprint]{acl}

\usepackage{times}
\usepackage{latexsym}

\usepackage[T1]{fontenc}

\usepackage[utf8]{inputenc}

\usepackage{microtype}

\usepackage{inconsolata}

\usepackage{graphicx}

\usepackage[utf8]{inputenc} 
\usepackage[T1]{fontenc}    
\usepackage{url}            
\usepackage{booktabs}       
\usepackage{amsmath}
\usepackage{amsfonts}       
\usepackage{amssymb}
\usepackage{mathtools}      
\usepackage{nicefrac}       
\usepackage{microtype}      
\usepackage{graphicx}
\usepackage{float}          
\usepackage{stfloats}       
\usepackage{algorithm}
\usepackage{algpseudocode}
\usepackage{tabularx}       
\usepackage{multirow}       
\usepackage{longtable}      
\usepackage{rotating}       
\usepackage{xcolor}         
\usepackage{enumitem}       
\usepackage{tcolorbox}      
\tcbuselibrary{breakable, skins}
\usepackage{upquote}        
\usepackage{placeins}       
\usepackage{xurl}
\usepackage{hyperref}       
\usepackage[noabbrev]{cleveref} 

\usepackage{inconsolata}

\newcolumntype{L}{>{\raggedright\arraybackslash}X}

\algrenewcommand\algorithmicrequire{\textbf{Input:}}
\algrenewcommand\algorithmicensure{\textbf{Output:}}
\algrenewcommand\algorithmiccomment[1]{\hfill\(\triangleright\)~\textit{#1}}

\crefname{algorithm}{Algorithm}{Algorithms}
\Crefname{algorithm}{Algorithm}{Algorithms}
\crefname{table}{Table}{Tables}
\Crefname{table}{Table}{Tables}
\crefname{figure}{Figure}{Figures}
\Crefname{figure}{Figure}{Figures}
\crefname{section}{Section}{Sections}
\Crefname{section}{Section}{Sections}
\crefname{subsection}{Section}{Sections}
\Crefname{subsection}{Section}{Sections}
\crefname{subsubsection}{Section}{Sections}
\Crefname{subsubsection}{Section}{Sections}
\crefname{appendix}{Appendix}{Appendices}
\Crefname{appendix}{Appendix}{Appendices}
\crefname{equation}{Eq.}{Eqs.}
\Crefname{equation}{Equation}{Equations}

\renewcommand{\dbltopfraction}{0.95}

\title{OracleProto: A Reproducible Framework for Benchmarking LLM Forecasting via Knowledge Cutoffs and Temporal Masking}

\author{
  \textbf{Yiding Ma}\textsuperscript{*}
  \quad
  \textbf{Chengyun Ruan}\textsuperscript{*}
  \quad
  \textbf{Kaibo Huang}\textsuperscript{\textdagger}
\\
  \textbf{Jin Tan}
  \quad
  \textbf{Zhongliang Yang}\textsuperscript{\textdagger}
  \quad
  \textbf{Linna Zhou}
\\[4pt]
  Beijing University of Posts and Telecommunications
\\[2pt]
  {\small\texttt{
    \{yidingma, ruanchengyun815, huangkaibo, tanjinanin, yangzl, zhoulinna\}@bupt.edu.cn
  }}
\\[4pt]
  {\small
    \textsuperscript{*}Equal contribution.
    \qquad
    \textsuperscript{\textdagger}Corresponding authors.
  }
}

\begin{document}
\maketitle

\begin{abstract}
Forecasting is a composite capability encompassing information gathering, evidence integration, situational assessment, and decision-making. Large language models (LLMs) combine reasoning and tool use, offering new avenues for automated forecasting and decision support. Yet LLM forecasting evaluation faces a structural challenge: live benchmarks support prospective assessment, but limited question reuse and changing information conditions make results difficult to reproduce and compare; retrospective benchmarks offer reproducibility but risk information leakage. We introduce \textbf{OracleProto}, a reproducible framework for evaluating LLM forecasting in a controlled, temporally masked environment. Through model knowledge cutoff constraints and temporal masking, OracleProto constructs a temporal sandbox for real-world web retrieval, audits information boundaries through leakage detection, and reconstructs resolved events as forecasting instances. Discrete answer normalization and hierarchical scoring enable unified evaluation. We evaluate 27 model--configuration pairs on 300 FutureX-Past-derived questions over 24,300 prediction trials, assessing forecasting quality, sampling stability, and cost efficiency under controlled information boundaries. An audit finds only 1.37\% residual leakage in retrieval results after temporal masking and leakage filtering. OracleProto retains resolved forecasting tasks as reusable evaluation resources under explicit temporal boundaries, supporting continual accumulation of forecasting datasets and systematic evaluation of forecasting capabilities. Code is available at \url{https://anonymous.4open.science/r/OracleProto-2E3A/}.
\end{abstract}

\begin{figure}[!t]
    \centering
    \includegraphics[width=\columnwidth]{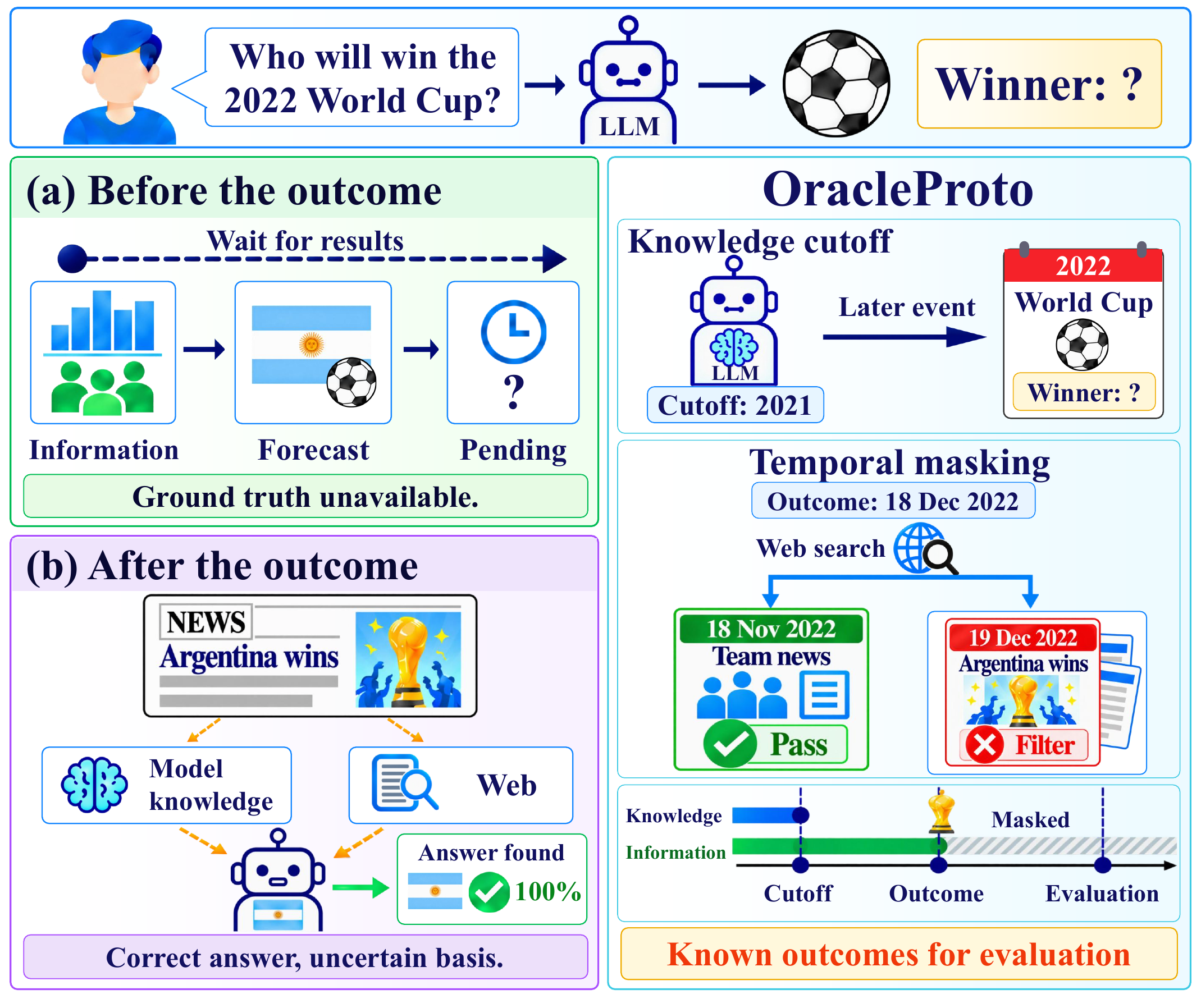}
    \caption{Forecasting evaluation challenges and OracleProto's knowledge cutoffs and temporal masking.}
    \label{fig:intro-motivation}
\end{figure}

\section{Introduction}
Forecasting helps people anticipate change, informing everyday risk responses and planning in science and technology. Large language models (LLMs) are increasingly used in decision-support settings~\citep{liu2024dellmadecisionmakinguncertainty,li-etal-2025-investorbench,goh2025gpt4,jain-etal-2026-decisive}. Forecasting requires models to connect information search, evidence integration, situational assessment, and decision-making into a complete reasoning chain when evidence is incomplete and evolving and outcomes remain unknown, rather than merely recalling facts. By combining reasoning with tool use, LLMs offer new avenues for automated forecasting, making forecasting a valuable setting for assessing evidence integration and reasoning under uncertainty. Their demonstrated potential to approach competitive human forecasters further motivates systematic evaluation~\citep{halawi2024approaching}.

However, evaluating LLM forecasting capabilities presents a structural challenge. ForecastBench and FutureX support prospective forecasting evaluation by continually updating questions about events whose outcomes remain unknown~\citep{karger2025forecastbench,zeng2025futurexadvancedlivebenchmark}. This approach reduces the risk of contamination from known answers, but the resolution of questions and changes in web pages and search results limit repeated evaluation on the same tasks, making score changes across evaluation rounds difficult to attribute directly to changes in model capabilities~\citep{chen-etal-2026-browsecomp}. Meanwhile, building and updating benchmarks requires effort in design, screening, and review~\citep{zhu-etal-2024-inference,wu-etal-2025-antileakbench}. If these high-value questions cannot be reused under controlled information conditions after their outcomes become known, they become single-use items, wasting the resources spent on dataset construction and annotation. Although retrospective evaluation allows scoring against known outcomes, models may access post-event information through prior knowledge or web retrieval, causing factual recall to be mistaken for forecasting ability~\citep{paleka2025pitfalls}. Prior studies show that specifying a historical date or asking models to ignore known outcomes in prompts cannot reliably restore their knowledge state at that time~\citep{liu-etal-2026-exante,li2026simulated}; search-engine date filters may likewise return pages containing later information, inflating apparent forecasting performance~\citep{el-lahib-etal-2026-temporal}. Offline web snapshots or fixed retrieval corpora can improve reproducibility~\citep{chen-etal-2026-browsecomp}, but their information coverage and update mechanisms differ from those of the open web, making it difficult to combine reproducible evaluation with the realism of live web retrieval~\citep{wildman2025btf,chai2026timemachine}.

\textbf{Whether resolved events can become forecasting tasks again depends on whether the evaluation system can re-establish clear temporal boundaries around the information available to the model.}

We therefore introduce \textbf{OracleProto}, a reproducible dataset-level framework for evaluating LLM forecasting in a controlled, temporally masked environment. OracleProto determines eligible historical questions from model knowledge cutoffs and event times, constrains external retrieval through temporal masking, and incorporates content-level leakage detection to construct a temporal sandbox with auditable information boundaries for real-world web retrieval. Resolved events are thus reconstructed as controlled forecasting instances, with known answers used only for scoring. Discrete answer normalization and hierarchical scoring further unify sample construction, controlled forecasting, and outcome evaluation, supporting model comparison under explicitly documented evaluation conditions.

Our evaluation combines information-boundary comparisons, model comparisons, and repeated trials to examine score sensitivity to retrieval and filtering conditions, alongside relationships among error patterns, reasoning configurations, and prediction consistency. Models with similar overall scores can exhibit different error patterns; greater reasoning effort does not necessarily yield consistent gains; and prediction consistency does not imply correctness. Understanding forecasting performance therefore requires examining the information conditions and model behavior behind scores. By retaining eligible historical tasks as reusable evaluation resources, OracleProto supports continual dataset accumulation and systematic forecasting evaluation. Our contributions are as follows:

\begin{itemize}[leftmargin=*]

  \item \textbf{A formal dataset-level framework for LLM forecasting evaluation.}
  We reconstruct the evaluation of LLM forecasting capability from one-time outcomes dependent on live events into dataset-level tasks that are definable, auditable, and reproducible. The framework provides not only a dataset, but also a continuously extensible method for sample construction.

  \item \textbf{The OracleProto evaluation protocol.}
  The protocol combines knowledge cutoffs, temporal masking, leakage detection, answer normalization, and hierarchical scoring to turn resolved events into controlled forecasting tasks. It exposes the same information boundaries and the same scoring interface to every model under test, so cross-model differences in measured accuracy reflect model behavior rather than shifting evaluation conditions.

  \item \textbf{A systematic evaluation benchmark and a forecasting evaluation framework.}
  Based on FutureX-Past~\citep{futurexpast2025}, we construct an evaluation set of 300 questions and evaluate 27 contemporary model--configuration pairs, with three trials per question, yielding 24,300 prediction trials. Experiments show that the framework can systematically assess the forecasting performance of different models across four dimensions: forecasting quality, sampling stability, cost efficiency, and residual leakage risk. This assessment reveals trade-offs among forecasting quality, stability, and cost, and identifies residual leakage risks that may affect the credibility of the evaluation.

\end{itemize}

\section{Related Work}

\label{sec:related-work}

\subsection{Large Language Models for Event Forecasting}

\label{sec:related-applications}

Prior work applies retrieval, tool use, and prediction aggregation to event forecasting with large language models. Halawi et al. combine information retrieval, forecast generation, and aggregation, with their system approaching the crowd aggregate of competitive human forecasters on their test set~\citep{halawi2024approaching}; MIRAI provides agents with structured events, news, and tool interfaces to evaluate international event forecasting~\citep{ye2024mirai}. BLF organizes iterative retrieval around a structured linguistic belief state and studies the roles of multi-trial aggregation and post-hoc calibration in binary probabilistic forecasting~\citep{murphy2026agentic}. Training methods also seek to improve forecasting: Time-R1 uses a reinforcement learning curriculum to develop temporal understanding and event prediction, while outcome-based reinforcement learning uses prediction-market questions and outcomes to optimize probabilistic forecasts~\citep{liu2025timer1,turtel2025outcomebased}. OpenForecaster uses an offline news corpus to construct forecasting tasks for training and support retrieval~\citep{chandak2026scaling}; Echo's official technical blog describes dynamic task generation and training with process-based rewards for unresolved events~\citep{unipat2026echo}.

\subsection{Forecasting Evaluation and Temporal Information Boundaries}

\label{sec:related-methodological}

Evaluation approaches differ in how they handle the timing of information availability. ForecastBench and FutureX continually collect questions about unresolved events for prospective evaluation~\citep{karger2025forecastbench,zeng2025futurexadvancedlivebenchmark}; Daily Oracle generates question--answer pairs from daily news to continuously assess temporal generalization and forecasting performance~\citep{dai2024prescient}. Historical backtesting enables scoring directly against known outcomes, but retaining questions and answers does not itself ensure valid forecasting evaluation: the FutureX-Past dataset documentation warns that newer models may already know historical outcomes~\citep{futurexpast2025}. BTF supports repeatable pastcasting through pre-collected web snapshots and RetroSearch, updating eligible questions as model knowledge cutoffs advance~\citep{wildman2025btf}. BLF combines model knowledge-cutoff criteria, search date restrictions, content-level leakage detection, and post-hoc auditing when evaluating historical questions~\citep{murphy2026agentic}.

The effectiveness of temporal constraints must still be assessed against leakage risks in models' existing knowledge and retrieved content. ExAnte and research on simulated ignorance show that specifying a historical date or instructing models to ignore known outcomes cannot reliably restore their historical knowledge state~\citep{liu-etal-2026-exante,li2026simulated}. Dated Data finds that effective knowledge cutoffs can vary across resources and topics, while LLMLagBench probes recent-event knowledge to identify temporal training boundaries~\citep{cheng2024dateddata,pezik2025llmlagbench}. Thus, disclosed knowledge cutoffs provide an operational basis for evaluation but do not guarantee that parametric knowledge excludes subsequent information. Research on forecasting evaluation pitfalls also highlights how temporal leakage and evaluation design affect conclusions about forecasting capability~\citep{paleka2025pitfalls}.

\begin{figure*}[t]
    \centering
    \includegraphics[width=\textwidth]{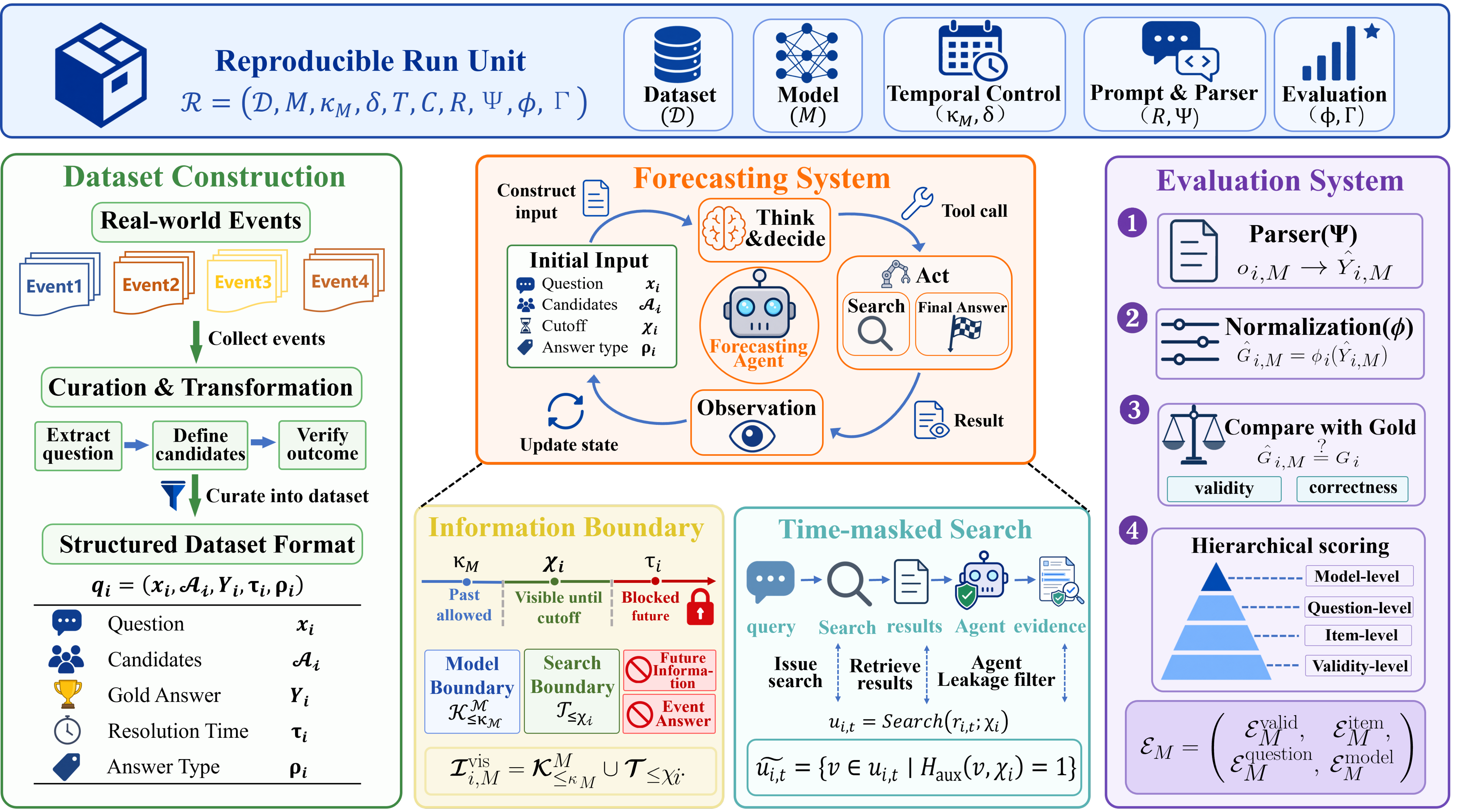}
    \caption{
    OracleProto overview. Resolved events are reconstructed as structured forecasting instances. Model knowledge cutoffs, temporal masking, and leakage filtering constrain the information available during forecasting. Predictions are parsed and normalized, then scored against verified outcomes at the item, question, and model levels.
    }
    \label{fig:figure1}
\end{figure*}

\section{Method}
\label{sec:method}

\subsection{Problem Formulation}

OracleProto evaluates LLM forecasting under a standardized bounded-retrieval harness: a fixed prompt renderer, search backend, leakage detector, parser, and scoring protocol that together expose the same information boundaries to every model under test. Each instance is a resolved event whose ground truth is hidden from the model and used only for scoring; the model must select from a finite candidate set using only information that would have been available before resolution. The model's knowledge cutoff constrains what may be encoded in parameters, and temporal masking constrains what may be retrieved during inference. The measured object is therefore the joint behavior of model and harness; isolating model-only forecasting capability would require ablating each harness component and lies outside the scope of this paper.

Let $\mathcal{D}=\{q_i\}_{i=1}^{N}$ with $q_i=(x_i,\mathcal{A}_i,Y_i,\tau_i,\rho_i)$, where $x_i$ is the question text, $\mathcal{A}_i$ is the finite candidate set, $Y_i\subseteq\mathcal{A}_i$ is the verified answer, $\tau_i$ is the resolution time, and $\rho_i$ specifies the selection structure (single- or multi-answer). The model receives only the time-masked view $q_i^{\mathrm{in}}=(x_i,\mathcal{A}_i,\chi_i,\rho_i)$, where $\chi_i$ is the prediction cutoff. For a model $M$ with knowledge cutoff $\kappa_M$, the admissible prediction set is
\begin{equation}
    \mathcal{D}^{\mathrm{pred}}_M=\{q_i\in\mathcal{D}\mid \kappa_M\le \chi_i<\tau_i\},
    \label{eq:pred-set}
\end{equation}
which excludes parametric leakage when $\chi_i<\kappa_M$ and post-resolution information when $\chi_i\ge\tau_i$. Instances violating this condition are logged with their exclusion reason rather than scored as forecasting failures, which is essential for comparing models with different cutoffs. The disclosed cutoff $\kappa_M$ is an operational proxy: effective cutoffs can drift from reported ones across resources and topics~\citep{cheng2024dateddata}, so $\kappa_M$ provides the strongest reproducible per-model boundary currently available rather than a parameter-level ground truth. The pipeline composes task construction, bounded inference, and scoring,
\begin{equation}
\mathcal{D}\to\mathcal{D}^{\mathrm{pred}}_M \to F_M(q_i^{\mathrm{in}};\mathcal{I}_{i,M}^{\mathrm{vis}}) \to\widehat{Y}_{i,M}\to\mathcal{E}_M,
\label{eq:pipeline}
\end{equation}
with $F_M$ the time-masked forecasting system, $\mathcal{I}_{i,M}^{\mathrm{vis}}$ the information visible to the model, and $\mathcal{E}_M$ the evaluation system. The time-replay assumption that licenses replaying resolved events as forecasting tasks is discussed in \cref{app:method-replay}.

\subsection{Forecasting Dataset Framework}

Each instance has a finite answer space ($|\mathcal{A}_i|\ge 2$), a verified answer subset $Y_i\subseteq\mathcal{A}_i$, and a resolution time $\tau_i$. Single-answer questions satisfy $|Y_i|=1$ and multi-answer questions satisfy $|Y_i|\ge 1$; the structural constraint $\rho_i$ determines the valid output cardinality. Given a temporal masking offset $\delta$, the prediction cutoff $\chi_i=\tau_i-\delta$ keeps the visible retrieval environment before the day on which the answer becomes verifiable. To remove surface variation in natural-language outputs, an answer normalization map $\phi_i:\mathcal{A}_i\to\mathcal{L}_i$ projects predictions and ground truth into a finite label set, so scoring compares $G_i=\phi_i(Y_i)$ and $\widehat{G}_{i,M}=\phi_i(\widehat{Y}_{i,M})$ as finite sets. Field-format details and the four resulting reproducibility properties are in \cref{app:dataset}.

\subsection{Time-Masked Forecasting System}

The forecasting system $F_M:(q_i^{\mathrm{in}},\mathcal{I}_{i,M}^{\mathrm{vis}})\to\widehat{Y}_{i,M}\subseteq\mathcal{A}_i$ returns a subset of the candidate answer set; no continuous score or subjective confidence is required. The visible information combines parametric knowledge before $\kappa_M$ with retrieval results before $\chi_i$,
\begin{equation}
    \mathcal{I}_{i,M}^{\mathrm{vis}}=\mathcal{K}^{M}_{\le\kappa_M}\cup\mathcal{T}_{\le\chi_i}.
    \label{eq:visible-info}
\end{equation}
The model's action space contains query proposal and answering; the cutoff $\chi_i$ is held by the evaluation system and injected at the tool layer, so the model cannot modify the information boundary.

Temporal masking enforces $\mathcal{T}_{\le\chi_i}$ at two layers. Given a model-proposed query $r_{i,t}$ at step $t$, the tool layer executes $u_{i,t}=\operatorname{Search}(r_{i,t};\chi_i)$, returning the snippet list through a date-restricted retrieval backend; our implementation uses Tavily, but the framework is backend-agnostic when the boundary is enforced by the evaluator. Returned snippets, cached pages, and aggregate summaries can still carry information after $\chi_i$, so an auxiliary detector $H_{\mathrm{aux}}$ performs a content-level audit before results enter the main-model context, yielding $\widetilde{u}_{i,t}=\{v\in u_{i,t}\mid H_{\mathrm{aux}}(v,\chi_i)=1\}$. $H_{\mathrm{aux}}$ sees only the retrieval payload (title, URL, publication time, snippet, cutoff date) and never the question, candidates, ground truth, or main-model dialogue, so it cannot serve as an answer auditor. Detection outcomes, latency, and error types are logged as audit metadata.

\paragraph{Forecasting interaction.}
For an admissible instance, the deterministic renderer $R$ produces the initial message $m_{i,0}=R(q_i^{\mathrm{in}})$, which contains the question, candidate set, prediction cutoff, selection structure, and the requirement that the final answer be parseable as a discrete prediction. The interaction is bounded by step cap $T$ and search-call cap $C$. At step $t$, conditioning only on the previous history, the model selects an action $a_{i,t}\in\mathcal{U}\cup\{\operatorname{answer}\}$, where $\mathcal{U}$ collects allowed tool actions and the key tool action is time-masked search. Filtered results are appended to the context as $m_{i,t}=m_{i,t-1}\oplus(a_{i,t},\widetilde{u}_{i,t})$ under deterministic concatenation $\oplus$; if a search is requested after the budget is exhausted, no external call is made and a budget-status message is appended so the model must answer within the remaining steps. When the model submits a final answer $o_{i,t}$, the parser $\widehat{Y}_{i,M}=\Psi_i(o_{i,t})$ checks $\mathcal{A}_i$ and $\rho_i$; missing answers, invalid formats, candidate-set mismatches, and structural-constraint violations are marked as invalid outputs, as is the case when no parseable answer is produced within $T$ steps. The full procedure is summarized in \cref{alg:time-masked-loop}.

\begin{algorithm}[t]
    \caption{Time-masked discrete forecasting loop. The cutoff is held by the evaluation system and injected at the tool layer; the evaluated model can observe questions, candidate answers, and filtered retrieval results, but cannot modify the temporal boundary.}
    \label{alg:time-masked-loop}
    \small
    \begin{algorithmic}[1]
        \Require Instance $q_i=(x_i,\mathcal{A}_i,Y_i,\tau_i,\rho_i)$, model $M$, knowledge cutoff $\kappa_M$, prediction cutoff $\chi_i$, maximum steps $T$, maximum search calls $C$
        \Ensure Parsed discrete prediction or a boundary/invalid-output marker

        \If{$\neg(\kappa_M \le \chi_i < \tau_i)$}
        \State \Return outside-boundary marker
        \EndIf

        \State $q_i^{\mathrm{in}} \gets (x_i,\mathcal{A}_i,\chi_i,\rho_i)$
        \State $m_0 \gets R(q_i^{\mathrm{in}})$
        \State $c \gets 0$

        \For{$t=1,\ldots,T$}
        \State $a_t \gets M(m_{t-1})$

        \If{$a_t$ is a final answer}
        \State \Return $\Psi_i(a_t)$
        \EndIf

        \If{$a_t$ is a search action and $c<C$}
        \State Let $r_t$ be the query proposed by $M$
        \State $u_t \gets \operatorname{Search}(r_t;\chi_i)$
        \Comment{tool-level date filtering}
        \State $\widetilde{u}_t \gets \operatorname{AuxLeakFilter}(u_t,\chi_i)$
        \Comment{content-level leakage detection}
        \State $m_t \gets m_{t-1}\oplus(a_t,\widetilde{u}_t)$
        \State $c \gets c+1$
        \Else
        \State $m_t \gets m_{t-1}\oplus \text{budget/status message}$
        \EndIf
        \EndFor

        \State \Return invalid-output marker
    \end{algorithmic}
\end{algorithm}

\subsection{Evaluation System}

Evaluation maps raw outputs $o_{i,M}$ through a parser $\Psi_i$ and the normalization $\phi_i$,
\begin{equation}
    o_{i,M}\xrightarrow{\Psi_i}\Psi_i(o_{i,M})\in 2^{\mathcal{A}_i}\cup\{\bot\}\xrightarrow{\phi_i}\widehat{G}_{i,M},
\end{equation}
where $\bot$ marks unevaluable outputs (missing answer, refusal, invalid format, or candidate-set mismatch). Validity and correctness are kept distinct: the evaluability indicator $v_{i,M}=\mathbb{I}[\Psi_i(o_{i,M})\ne\bot]$ asks whether the output is parseable, while the strict item-level correctness indicator $r_{i,M}=\mathbb{I}[\widehat{G}_{i,M}=G_i]$ (defined when $v_{i,M}=1$) asks whether the normalized prediction equals the gold set. The full evaluation system has four levels,
\begin{equation}
    \mathcal{E}_M=(\mathcal{E}^{\mathrm{valid}}_M,\mathcal{E}^{\mathrm{item}}_M,\mathcal{E}^{\mathrm{question}}_M,\mathcal{E}^{\mathrm{model}}_M),
\end{equation}
measuring parseability, single-prediction correctness, repeated-trial behavior on one question, and aggregate behavior on $\mathcal{D}^{\mathrm{pred}}_M$, respectively. If a question is run independently $S$ times, the question-level object $\{\widehat{G}_{i,M}^{(s)}\}_{s=1}^{S}$ captures accuracy, stability, consistency, and coverage across repeated attempts. At the model level, the aggregator $\Gamma$ collapses $\{\mathcal{E}^{\mathrm{question}}_{i,M}\}_{q_i\in\mathcal{D}^{\mathrm{pred}}_M}$ to a single scalar and may report accuracy, evaluability rate, stability across repeated trials, consistency across question types, multi-answer handling, or tool-use cost. Multi-answer set-difference relations and the full evaluation chain are in \cref{app:method-eval}.

\begin{figure*}[!b]
 \centering
 \includegraphics[width=\textwidth]{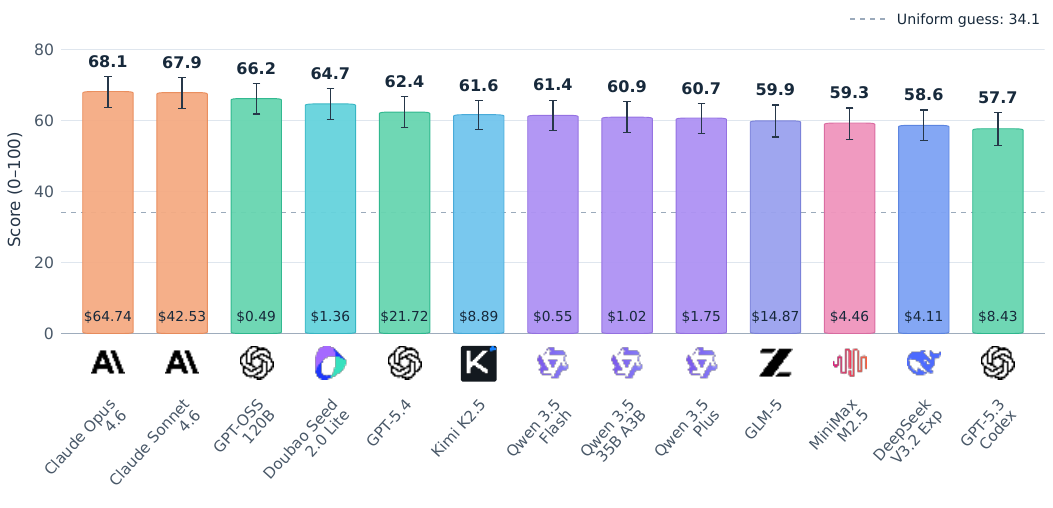}
 \caption{Scores for 13 default configurations on 300 questions, with $95\%$ confidence intervals. Bar labels estimate forecasting-model cost (USD per 300-question trial) at fixed public tariffs. All 27 configurations appear in \cref{tab:complete-results}.}
 \label{fig:score-bar}
\end{figure*}

\subsection{Reproducibility and Leakage Boundary}

A complete run is the tuple
\begin{equation}
    \mathcal{R}=(\mathcal{D},M,\kappa_M,\delta,T,C,R,\Psi,\phi,\Gamma).
\end{equation}
Once $\mathcal{R}$ is fixed, sample admission, prompt rendering, tool masking, parsing, and aggregation have an auditable replay path, and per-question stochasticity is captured by recording the sampling settings and the index $s$ of each independent run. The detector $H_{\mathrm{aux}}$, its prompt SHA-256, and its failure policy enter $\mathcal{R}$ as run-configuration metadata, so the leakage barrier is itself byte-reproducible.

OracleProto controls leakage through three mechanisms: \emph{parametric knowledge} via admissibility (\cref{eq:pred-set}), \emph{tool-mediated knowledge} via the cutoff $\chi_i$ injected at the tool layer (\cref{eq:visible-info}), and \emph{retrieval-result content} via $H_{\mathrm{aux}}$. Residual risks not closed by these three are cataloged in \cref{app:method-channels}.

\input{experiments.tex}

\section{Conclusion}
\label{sec:conclusion}

We introduce OracleProto, a reproducible dataset construction paradigm and evaluation framework for assessing LLM forecasting capabilities. By unifying sample construction, controlled forecasting, and outcome evaluation, OracleProto reconstructs resolved historical events as forecasting instances with explicit temporal boundaries, enabling historical forecasting tasks to be reused and evaluated under a specified protocol. Experiments on FutureX-Past demonstrate its application to evaluation set construction and cross-model assessment, providing a systematic foundation for building and extending forecasting benchmarks. Future work will expand dataset size and domain coverage and investigate the sources of model performance differences through more detailed analysis of forecasting processes.

\section*{Limitations}

OracleProto relies on publicly disclosed model knowledge cutoffs and temporal masking mechanisms to construct controlled information boundaries, while the actual training data of deployed models and the web retrieval environment are not always fully observable. A small amount of undetected residual leakage may therefore remain. The experiments use 300 discrete-choice questions derived from FutureX-Past together with a fixed retrieval backend, search budget, and scoring protocol. These choices support auditability, but the current results mainly reflect model performance in controlled discrete forecasting settings rather than open-ended forecasting, continuous probabilistic forecasting, or longer-horizon real-world applications.

\bibliography{custom}

\clearpage
\appendix

\input{appendix_experiments.tex}

\end{document}

%% file: experiments.tex
\section{Experiment}
\label{sec:experiment}
\input{experiment_assets/chapter4_numbers.tex}
\providecommand{\cmark}{{\usefont{U}{pzd}{m}{n}\char51}}
\providecommand{\pmark}{{\usefont{U}{pzd}{m}{n}\char119}}
\providecommand{\xmark}{{\usefont{U}{pzd}{m}{n}\char55}}

\label{sec:setup}
\paragraph{Setup.}
The main benchmark evaluates 27 model configurations on 300 discrete-choice questions from a FutureX-Past-derived pool~\citep{futurexpast2025,zeng2025futurexadvancedlivebenchmark}: 109 binary, 154 multiple-choice single-answer (MCS), and 37 multiple-answer (MM), with three trials each (24,300 predictions). A trial allows six model steps, four Tavily searches of five results, and 12,000 completion tokens per call; the inclusive retrieval cutoff is the day before resolution (\cref{app:dataset,app:configuration}).

\begin{table}[t]
 \centering\footnotesize
 \setlength{\tabcolsep}{3.2pt}
 \renewcommand{\arraystretch}{1.05}
 \begin{tabular}{@{}lcccccc@{}}
 \toprule
 & Live & Reuse & Web & Snap. & Cutoff & Leak \\
 \midrule
 ForecastBench & \cmark & \xmark & \pmark & \xmark & -- & -- \\
 FutureX & \cmark & \xmark & \cmark & \xmark & -- & -- \\
 FutureX-Past & \xmark & \xmark & \xmark & \xmark & \xmark & \xmark \\
 \midrule
 BTF & \xmark & \pmark & \pmark & \cmark & \pmark & \xmark \\
 FutureSim & \xmark & \pmark & \xmark & \cmark & \pmark & \xmark \\
 \midrule
 Halawi et al. & \xmark & \pmark & \pmark & \xmark & \pmark & \xmark \\
 AIA Forecaster & \pmark & \pmark & \cmark & \xmark & \pmark & \pmark \\
 BLF & \xmark & \pmark & \cmark & \xmark & \pmark & \pmark \\
 Time Machine & \xmark & \pmark & \cmark & \xmark & \xmark & \pmark \\
 \midrule
 OracleProto & \xmark & \cmark & \cmark & \pmark & \cmark & \cmark \\
 \bottomrule
 \end{tabular}
 \caption{Positioning: \cmark{}~full, \pmark{}~partial, \xmark{}~absent, --~n/a. Full Reuse: valid for newer models without new data; Cutoff: per-question admission; Leak: filtering evidence before the model reads it, plus a human-verified audit.}
 \label{tab:positioning}
\end{table}

\label{sec:metrics}
\paragraph{Score.}
Following FutureX's choice-question metrics~\citep{zeng2025futurexadvancedlivebenchmark}, trial~$r$ of question~$i$ earns exact-match credit $c_{ir}=\mathbb{I}[\widehat G_{ir}=G_i]$ on binary and MCS and set F1 $c_{ir}=2\mathrm{TP}/(2\mathrm{TP}+\mathrm{FP}+\mathrm{FN})$ on MM, with TP, FP, and FN counting correct, extra, and missed options; mirroring its heavier weighting of harder tiers, the types weigh $w_i=1$, 2, and 3 points (528 in total). Over question set $\mathcal Q$, $\mathrm{Score}=100\sum_{i\in\mathcal Q}w_i\bar c_i/\sum_{i\in\mathcal Q}w_i$, with $\bar c_i$ the mean of $c_{ir}$ over three trials. Refusals and invalid answers earn zero. Intervals use a paired question-cluster bootstrap stratified by type (\cref{app:significance}).

\label{sec:leakage-audit}

\begin{table*}[t]
 \centering\footnotesize
 \setlength{\tabcolsep}{2pt}
 \renewcommand{\arraystretch}{1.08}
 \input{experiment_assets/tables/boundary_comparison.tex}
 \caption{Boundary conditions on \BoundaryQuestionCount{} questions, \BoundaryTrialCount{} trials per condition. $N$: no retrieval; $F$: date restriction plus content filtering; $D$: date restriction only. Differences in Score with unadjusted 95\% paired bootstrap CIs. Evidence retained: share of result occurrences kept in the same saved candidate batches before adaptive query selection.}
 \label{tab:boundary-comparison}
\end{table*}

\paragraph{Protocol positioning.}
Prior designs secure replayability or the information boundary, not both (\cref{tab:positioning}). FutureX rejects backtesting as past-event search results reflect later knowledge~\citep{zeng2025futurexadvancedlivebenchmark}. Snapshot replays freeze evidence at crawl time, so newer cutoffs need new questions and crawls~\citep{wildman2025btf,goel2026futuresim}. Live-search backtests keep the open web but bound parametric knowledge only by test window or model choice, if at all~\citep{halawi2024approaching,alur2025aia,murphy2026agentic,chai2026timemachine}; their leakage checks rely on LLM verdicts alone (BLF), run after forecasting (AIA Forecaster), or read the gold answer, possibly discarding valid evidence pointing to it, with no human labels reported (Agentic Time Machine). OracleProto admits questions per model and filters date-restricted live results blind to question and answer, keeping a resolved question evaluable for any model whose cutoff precedes it; as live search is not byte-stable, it archives queries and payloads, not a frozen corpus.

\paragraph{Controlled comparison.}
Separately from the main benchmark, GLM-5, Kimi K2.5, Doubao Seed 2.0 Lite, and Qwen 3.5 Flash answer the same \BoundaryQuestionCount{} questions (\BoundaryPredictionCount{} predictions) under conditions varying the Web and Leak columns of \cref{tab:positioning}: no retrieval ($N$); date restriction ($D$), as in date-filtered backtests~\citep{paleka2025pitfalls,el-lahib-etal-2026-temporal}; and the full protocol ($F$), adding Qwen 3.8 Flash content filtering. Model identifiers, prompts, sampling settings, and budgets are fixed across conditions (OpenRouter, default provider routing). Both retrieval conditions replay a frozen bank of archived candidates per model, question, and trial: recorded queries return their saved batches, new queries are ranked lexically and deterministically within the bank, and $F$ filters selected candidates without replacement. This fixes the corpus, removing fresh-search variation, though received evidence may still steer queries.

\paragraph{Score sensitivity and evidence retention.}
Estimates follow $N<F<D$ for \BoundaryOrderedModels{} of the \BoundaryModelCount{} models (\cref{tab:boundary-comparison}): $F-N$, the gain from permitted retrieval evidence, spans $\BoundaryRetrievalMin$ to $\BoundaryRetrievalMax$ Score points, and $D-F$, the sensitivity to removing flagged candidates, $\BoundaryFilterMin$ to $\BoundaryFilterMax$. Even the smallest $D-F$ exceeds the 11.0-point range of all 27 configurations (\cref{tab:complete-results}), so an unreported boundary can outweigh ranked model differences, and a higher date-only Score alone cannot separate legitimate evidence gains from access to realized outcomes. Filtering retains \BoundaryRetainedMin--\BoundaryRetainedMax\% of candidate result occurrences, measuring evidence availability, not useful information. As the filter also drops future-dated schedules and speculation the audit permits (\cref{app:leakage}), and removed results may mix post-cutoff facts with admissible context, $D-F$ is not a pure estimate of leakage-induced inflation.

\paragraph{Residual leakage.}
In a separate audit of the main benchmark's post-filter evidence, GPT 6.1 Sol flags \AuditLeakCount{} of \AuditResultCount{} results (\AuditLeakPercent\%) as containing post-cutoff realized facts; human verification confirms all \AuditHumanPositive{} positives and finds no leakage in \AuditHumanNegative{} sampled negatives (\cref{app:leakage}). This rate describes the audited post-filter population, not a paired estimate of leakage reduction in the ablation.

\label{sec:results}

\begin{figure}[t]
 \centering
 \includegraphics[width=\columnwidth,trim=0 5 0 2.5,clip]{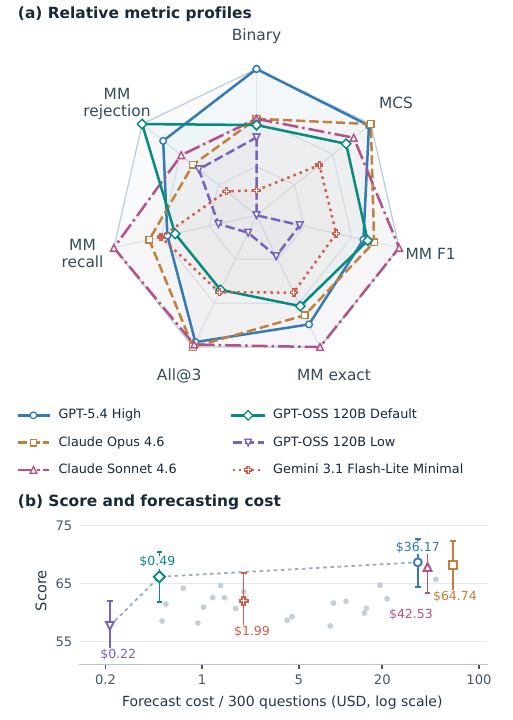}
 \caption{Task profiles and cost of six representative configurations. Each radar axis is min--max scaled over all 27 configurations (outward is higher). Cost: log scale, 95\% Score CIs, dashed point-estimate frontier, others in gray. Raw metrics, axis ranges: \cref{app:profile-selection}.}
 \label{fig:answer-profile}
\end{figure}

\paragraph{Similar totals conceal different error patterns.}
Claude Opus 4.6 leads Claude Sonnet 4.6 by only $\OpusSonnetDelta$ Score points (95\% CI $\OpusSonnetCI$; \cref{tab:paired-deltas}), yet Opus has higher MCS accuracy and Sonnet higher MM credit and exact accuracy (\cref{fig:answer-profile}); the aggregate weights these strengths by question type rather than identifying a uniformly stronger model. In MM, GPT-OSS 120B Default rejects incorrect options best but misses more correct ones than Sonnet, trading recovery for fewer additions, and Sonnet's \SonnetMMCredit\% credit versus \SonnetMMExact\% exact accuracy shows that partial recovery is not complete correctness (\cref{app:mm-diagnostics,app:cases}).

\begin{table}[t]
 \centering\footnotesize
 \setlength{\tabcolsep}{1pt}
 \renewcommand{\arraystretch}{1.0}
 \input{experiment_assets/tables/paired_deltas.tex}
 \caption{Selected comparisons between and within models. $\Delta$ (Score): A minus B, or evaluated minus reference setting; Opus and Sonnet use Default; CIs unadjusted.}
 \label{tab:paired-deltas}
\end{table}

\paragraph{Cost adds a separate decision criterion.}
GPT-OSS Default trails Opus by $\OpusOSSDelta$ Score points but costs \$\OSSCost{} versus \$\OpusCost{} per 300-question trial at recorded tariffs, without fewer searches or model steps (\cref{tab:resources}). As leading configurations' paired CIs include zero (\cref{tab:paired-deltas}), read the cost frontier with uncertainty and error profiles (\cref{app:cost}).

\label{sec:reasoning}

\begin{figure}[t]
 \centering
 \includegraphics[width=\columnwidth,trim=0 4.5 0 3.5,clip]{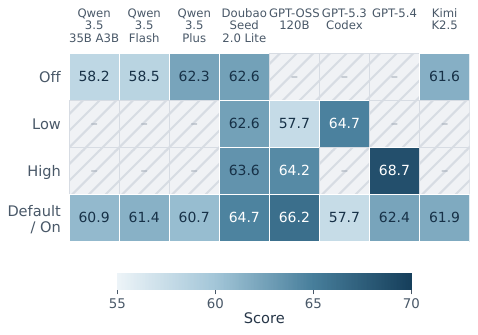}
 \caption{Score across provider-specific reasoning settings; hatched cells were not evaluated.}
 \label{fig:reasoning-heatmap}
\end{figure}

\paragraph{More reasoning need not help.}
Effects vary by model (\cref{fig:reasoning-heatmap,tab:paired-deltas}): GPT-5.4 High beats Default by $\GPTReasoningDelta$ points ($\GPTReasoningCI$), cutting MM omission from \GPTDefaultOmission\% to \GPTHighOmission\% and false selection from \GPTDefaultFalseSelection\% to \GPTHighFalseSelection\%. Kimi's reasoning lowers false selection from \KimiOffFalseSelection\% to \KimiOnFalseSelection\% but raises omission from \KimiOffOmission\% to \KimiOnOmission\%, moving Score only $\KimiReasoningDelta$ ($\KimiReasoningCI$). Default even beats High for GPT-OSS, and Low beats Default for Codex; providers' settings share no common effort scale (\cref{app:configuration}).

\paragraph{Prediction consistency is not correctness.}
Strict majority voting over complete answer sets scores below the trial mean for \VoteWorseCount{} of \MainConfigurationCount{} configurations: partly correct MM sets may lack a majority, while agreeing on a wrong set keeps the error. Opus is exact in all trials on \OpusExactAll\% of questions but repeats a wrong set on \OpusSameWrong\%; GPT-OSS Default, scoring similarly, has \OSSExactAll\% and \OSSSameWrong\% (\cref{app:repetition}).

%% file: experiment_assets/chapter4_numbers.tex
\newcommand{\BoundaryQuestionCount}{300}
\newcommand{\BoundaryTrialCount}{3}
\newcommand{\BoundaryPredictionCount}{10,800}
\newcommand{\BoundaryFullPoints}{528}
\newcommand{\BoundaryModelCount}{4}
\newcommand{\BoundaryOrderedModels}{4}
\newcommand{\BoundaryRetrievalMin}{+10.65}
\newcommand{\BoundaryRetrievalMax}{+18.56}
\newcommand{\BoundaryFilterMin}{+12.65}
\newcommand{\BoundaryFilterMax}{+16.35}
\newcommand{\BoundaryRetainedMin}{40.4}
\newcommand{\BoundaryRetainedMax}{42.8}
\newcommand{\OpusOmission}{23.8}
\newcommand{\OpusFalseSelection}{20.2}
\newcommand{\OpusExactAll}{50.3}
\newcommand{\OpusSameWrong}{18.0}
\newcommand{\OpusCost}{64.74}
\newcommand{\OpusMMCredit}{74.3}
\newcommand{\OpusMMExact}{36.9}
\newcommand{\SonnetOmission}{18.1}
\newcommand{\SonnetFalseSelection}{18.3}
\newcommand{\SonnetExactAll}{50.0}
\newcommand{\SonnetSameWrong}{16.7}
\newcommand{\SonnetCost}{42.53}
\newcommand{\SonnetMMCredit}{78.3}
\newcommand{\SonnetMMExact}{43.2}
\newcommand{\OSSOmission}{28.0}
\newcommand{\OSSFalseSelection}{11.8}
\newcommand{\OSSExactAll}{42.7}
\newcommand{\OSSSameWrong}{10.7}
\newcommand{\OSSCost}{0.49}
\newcommand{\OSSMMCredit}{73.3}
\newcommand{\OSSMMExact}{35.1}
\newcommand{\GPTDefaultOmission}{35.2}
\newcommand{\GPTDefaultFalseSelection}{21.1}
\newcommand{\GPTDefaultExactAll}{40.7}
\newcommand{\GPTDefaultSameWrong}{13.0}
\newcommand{\GPTDefaultCost}{21.72}
\newcommand{\GPTDefaultMMCredit}{64.2}
\newcommand{\GPTDefaultMMExact}{26.1}
\newcommand{\GPTHighOmission}{26.8}
\newcommand{\GPTHighFalseSelection}{15.3}
\newcommand{\GPTHighExactAll}{49.7}
\newcommand{\GPTHighSameWrong}{12.0}
\newcommand{\GPTHighCost}{36.17}
\newcommand{\GPTHighMMCredit}{72.6}
\newcommand{\GPTHighMMExact}{38.7}
\newcommand{\KimiOffOmission}{30.3}
\newcommand{\KimiOffFalseSelection}{26.5}
\newcommand{\KimiOffExactAll}{37.7}
\newcommand{\KimiOffSameWrong}{10.3}
\newcommand{\KimiOffCost}{8.89}
\newcommand{\KimiOffMMCredit}{64.6}
\newcommand{\KimiOffMMExact}{25.2}
\newcommand{\KimiOnOmission}{32.9}
\newcommand{\KimiOnFalseSelection}{17.8}
\newcommand{\KimiOnExactAll}{42.0}
\newcommand{\KimiOnSameWrong}{15.0}
\newcommand{\KimiOnCost}{10.96}
\newcommand{\KimiOnMMCredit}{65.8}
\newcommand{\KimiOnMMExact}{29.7}
\newcommand{\OpusSonnetDelta}{+0.30}
\newcommand{\OpusSonnetCI}{[-2.68,\;+3.35]}
\newcommand{\GPTReasoningDelta}{+6.31}
\newcommand{\GPTReasoningCI}{[+2.50,\;+10.06]}
\newcommand{\KimiReasoningDelta}{+0.31}
\newcommand{\KimiReasoningCI}{[-2.88,\;+3.58]}
\newcommand{\OpusOSSDelta}{1.98}
\newcommand{\VoteWorseCount}{27}
\newcommand{\MainConfigurationCount}{27}
\newcommand{\AuditResultCount}{10,000}
\newcommand{\AuditLeakCount}{137}
\newcommand{\AuditHumanPositive}{137}
\newcommand{\AuditHumanNegative}{1,000}
\newcommand{\AuditLeakPercent}{1.37}

%% file: experiment_assets/tables/boundary_comparison.tex
\begin{tabular*}{\textwidth}{@{\extracolsep{\fill}}lrrrrrr@{}}
\toprule
& \multicolumn{3}{c}{Score} & \multicolumn{2}{c}{Paired difference [95\% CI]} & Evidence \\
\cmidrule(lr){2-4}\cmidrule(lr){5-6}
Model & None ($N$) & Full ($F$) & Date ($D$) & $F-N$ & $D-F$ & retained \\
\midrule
GLM-5 & 45.45 & 56.10 & 72.45 & $+10.65\;[+6.40,+15.01]$ & $+16.35\;[+12.55,+20.04]$ & 42.8\% \\
Kimi K2.5 & 41.45 & 58.09 & 73.21 & $+16.64\;[+11.52,+21.66]$ & $+15.12\;[+10.85,+19.15]$ & 40.5\% \\
Doubao Seed 2.0 Lite & 42.23 & 60.79 & 73.44 & $+18.56\;[+13.80,+23.20]$ & $+12.65\;[+8.97,+16.39]$ & 40.4\% \\
Qwen 3.5 Flash & 41.07 & 58.93 & 72.03 & $+17.85\;[+13.20,+22.35]$ & $+13.10\;[+9.59,+16.86]$ & 42.6\% \\
\bottomrule
\end{tabular*}

%% file: experiment_assets/tables/paired_deltas.tex
\begin{tabular*}{\columnwidth}{@{\extracolsep{\fill}}lllrr@{}}
\toprule
\multicolumn{3}{@{}l}{\textbf{Between models (A vs. B)}} & \textbf{$\Delta$} & \textbf{95\% CI} \\
\midrule
\multicolumn{3}{@{}l}{Opus 4.6 vs. Sonnet 4.6} & $+0.30$ & $[-2.68,\;3.35]$ \\
\multicolumn{3}{@{}l}{GPT-5.4 High vs. Opus 4.6} & $+0.53$ & $[-3.08,\;4.03]$ \\
\midrule
\textbf{Within model} & \textbf{Eval.} & \textbf{Ref.} & & \\
GPT-5.4 & High & Default & $+6.31$ & $[2.50,\;10.06]$ \\
GPT-5.3 Codex & Low & Default & $+7.06$ & $[3.55,\;10.84]$ \\
GPT-OSS 120B & High & Low & $+6.48$ & $[2.75,\;10.31]$ \\
Doubao Seed 2.0 Lite & High & Off & $+0.99$ & $[-1.97,\;4.03]$ \\
Qwen 3.5 Plus & On & Off & $-1.66$ & $[-4.80,\;1.37]$ \\
Qwen 3.5 Flash & On & Off & $+2.93$ & $[-0.31,\;6.23]$ \\
Qwen 3.5 35B A3B & On & Off & $+2.74$ & $[-0.91,\;6.51]$ \\
Kimi K2.5 & On & Off & $+0.31$ & $[-2.88,\;3.58]$ \\
\bottomrule
\end{tabular*}

%% file: appendix_experiments.tex
\raggedbottom
\setcounter{dbltopnumber}{1}
\renewcommand{\dbltopfraction}{0.72}
\captionsetup{skip=6pt}
\setlength{\floatsep}{10pt plus 2pt minus 2pt}
\setlength{\textfloatsep}{12pt plus 2pt minus 2pt}
\setlength{\intextsep}{10pt plus 2pt minus 2pt}
\setlength{\dblfloatsep}{12pt plus 2pt minus 2pt}
\setlength{\dbltextfloatsep}{12pt plus 2pt minus 2pt}
\makeatletter
\setlength{\@fptop}{0pt}
\setlength{\@fpsep}{12pt}
\setlength{\@fpbot}{0pt plus 1fil}
\setlength{\@dblfptop}{0pt}
\setlength{\@dblfpsep}{12pt}
\setlength{\@dblfpbot}{0pt plus 1fil}
\makeatother

\newcommand{\appendixdatatable}[1]{%
 \begingroup
 \renewcommand{\arraystretch}{1.0}
 \renewenvironment{tabular}[1]{%
  \begin{tabular*}{\linewidth}{@{\extracolsep{\fill}}##1}%
 }{\end{tabular*}}%
 #1%
 \endgroup
}
\newcommand{\appendixheader}[1]{%
 \let\appendixoriginaltoprule\toprule
 \def\toprule##1\\{\appendixoriginaltoprule#1\\}%
}
\newcommand{\pdef}{\textsuperscript{\dag}}
\newtcolorbox{apppanel}[2][]{enhanced, colback=gray!4, colframe=gray!55,
 boxrule=0.4pt, arc=1pt, left=4pt, right=4pt, top=3pt, bottom=3pt,
 toptitle=1.5pt, bottomtitle=1.5pt, colbacktitle=gray!14, coltitle=black,
 fonttitle=\small\bfseries, title={#2}, fontupper=\small,
 before skip=8pt, after skip=8pt, #1}
\newcommand{\casemeta}[3]{%
 \noindent\textit{Resolution:} #1;\enspace\textit{cutoff:} #2.\\
 \textit{Gold:} #3\par\smallskip}
\newenvironment{casetrials}{%
 \par\noindent\footnotesize\setlength{\tabcolsep}{3pt}%
 \tabularx{\linewidth}{@{}>{\raggedright\arraybackslash}Xllr@{}}
 \toprule
 Trial & Kept & Prediction & Credit\\
 \midrule
}{%
 \bottomrule
 \endtabularx\par
}

\section{Dataset Records and Time Boundaries}
\label{app:dataset}

The benchmark comprises 300 frozen question records derived from FutureX-Past~\citep{futurexpast2025}, with recorded resolution dates from March 12 to September 21, 2026. Each record stores an ID, the question text $x_i$, the ordered candidate options $\mathcal{A}_i$, the gold answer set $Y_i$, the selection type $\rho_i$ (single- or multiple-answer), the question type (yes/no, named binary, or multiple choice), and the resolution date $\tau_i$. Four properties make the records reproducible: (i)~\emph{fixed question identity}, as the ID joins each record to all of its trials; (ii)~\emph{a deterministic answer mapping}, as candidate order fixes the option letters; (iii)~\emph{an explicit time boundary}, as $\tau_i$ determines the retrieval cutoff (\cref{app:method-replay}); and (iv)~\emph{a fixed scoring denominator}, as every question keeps its weight and three trial slots whatever the model outputs (\cref{tab:question-distribution,app:method-eval}). The data manifest records the source-database hash and the scoring contract, and all 27 configurations share the same question IDs, answer mapping, item weights, and three-trial protocol.

\begin{table}[!htbp]
 \centering\small
 \begin{tabular}{@{}lrr@{}}
 \toprule
 Question type & Count & Points\\
 \midrule
 Yes/no & 100 & 1\\
 Named binary & 9 & 1\\
 MCS & 154 & 2\\
 MM & 37 & 3\\
 \midrule
 Total & 300 & 528\\
 \bottomrule
 \end{tabular}
 \caption{Question counts and scoring weights (528 points in total). Yes/no and named-binary questions form the 109 binary questions.}
 \label{tab:question-distribution}
\end{table}

\subsection{Time-Replay Assumption}
\label{app:method-replay}

Replaying a resolved event as a forecasting task assumes that the prediction cutoff $\chi_i$ recreates the information state before the outcome was known: the model's parametric knowledge ends by $\chi_i$, and retrieval exposes only information available by $\chi_i$. Admissibility (\cref{eq:pred-set}) enforces the first condition per model; temporal masking and leakage filtering enforce the second (\cref{app:routing,app:detector}).

The masking offset is $\delta=1$ day, so the inclusive retrieval cutoff is $\chi_i=\tau_i-1$ day: material dated $\chi_i$ may be retrieved, while material dated $\tau_i$ or later may not. Every configuration has a recorded admission date (\cref{tab:reasoning-semantics}); the latest, March 10, 2026, precedes the earliest cutoff, March 11, 2026, so all 300 questions are admissible for all 27 configurations.

Two kinds of items violate the assumption regardless of masking. First, the question stem or option order may presuppose the outcome, for example through post-resolution wording. Second, the outcome may be publicly reported on or before $\chi_i$ although the recorded resolution date is later; the Monaco item is an instance, with a race report dated on the permitted cutoff day (\cref{app:case-monaco}). An outcome that is settled before $\chi_i$ but not yet public, by contrast, remains a legitimate forecasting target. Removing items after inspecting model answers would change the scoring population, so the reported results keep the fixed 300-question set, including the Monaco item; such items are reported individually, and any reduced population would be a separate analysis.

\subsection{Record and Rendering Contract}
\label{app:record-contract}

The renderer reads the question text, resolution date, candidates, and a type-specific response specification; gold labels are read only by the offline scorer. Yes/no prompts request a boxed Yes or No, named-binary prompts one of the two outcome strings, MCS prompts one letter, and MM prompts a comma-separated set of letters. The resolution date in the prompt identifies the event instance and is never replaced by another reporting cycle, while the retrieval cutoff is supplied separately at the tool layer (\cref{app:routing}). Because candidate order defines the letter mapping, it must stay unchanged whenever prompts or scores are regenerated.

\section{Run Configuration}
\label{app:configuration}

\paragraph{Interaction budget.}
Each trial allows $T=6$ model steps, $C=4$ Tavily search calls of five results each, and 12,000 completion tokens per model call. A step may request several searches, and a model may answer before exhausting either budget (\cref{app:protocol}). The task template supplies the question, options, selection rule, resolution date, and boxed-answer format, and asks the model to decompose the question, search, cross-check sources, and commit to an answer (\cref{app:routing}).

\paragraph{Model panel.}
\Cref{tab:reasoning-semantics} lists the 27 configurations: 15 models, each run under one to four provider-specific reasoning settings. Each configuration has 900 protocol outcomes (300 questions, three trials), with refusals and invalid outputs scored zero (\cref{app:method-eval}). The 13 provider-default configurations form \cref{fig:score-bar}, and \cref{fig:reasoning-heatmap} groups the settings as Off, Low, High, and Default/On.

\paragraph{Sampling.}
Temperatures are listed in \cref{tab:reasoning-semantics}. Top-$p$ is 1.0 when supplied; some Claude requests omit it. Model profiles, sampling fields, filter versions, and admission dates are archived per source run in \texttt{source\_\allowbreak configuration.json}.

\begin{table*}[!tbp]
 \centering\small
 \begin{tabularx}{\textwidth}{@{}lLcc@{}}
 \toprule
 Model & Evaluated reasoning settings & Admission date & Temperature\\
 \midrule
 Claude Opus 4.6 & Default\pdef & 2026-02-04 & 1.0\\
 Claude Sonnet 4.6 & Default\pdef & 2026-02-17 & 1.0\\
 GPT-5.4 & Default\pdef; High (separate route) & 2026-03-05 & 0.7; High 1.0\\
 GPT-5.3 Codex & Default\pdef, Low & 2026-02-24 & 1.0\\
 GPT-OSS 120B & Default\pdef, Low, High & 2024-01-30 & 0.7\\
 Gemini 3.1 Pro (CT) & High & 2026-02-19 & 1.0\\
 Gemini 3.1 Flash-Lite & Minimal & 2026-03-03 & 1.0\\
 Qwen 3.5 Plus & On\pdef, Off & 2026-02-15 & 0.7\\
 Qwen 3.5 Flash & On\pdef, Off & 2026-02-25 & 0.7\\
 Qwen 3.5 35B A3B & On\pdef, Off & 2026-02-25 & 1.0\\
 Doubao Seed 2.0 Lite & Default\pdef, Off, Low, High & 2026-03-10 & 0.7\\
 Kimi K2.5 (AliCloud) & Off\pdef, On & 2026-01-27 & 0.7\\
 GLM-5 & Default\pdef, On & 2026-02-11 & 0.7\\
 MiniMax M2.5 & Default\pdef & 2026-02-12 & 0.7\\
 DeepSeek V3.2 Exp & Default\pdef & 2025-09-29 & 0.7\\
 \bottomrule
 \end{tabularx}
 \caption{The 27 evaluated configurations. \dag: provider-default setting, plotted in \cref{fig:score-bar}; the two Gemini models have no provider-default run. Admission date: the configured $\kappa_M$ archived with the run (\cref{app:configuration-provenance}). CT: custom-tools endpoint.}
 \label{tab:reasoning-semantics}
\end{table*}

\subsection{Model-Visible Prompt and Tool Boundary}
\label{app:routing}

The initial message combines a forecasting-role instruction, the question with its resolution date, the candidate list when applicable, and the type-specific output clause; budget and forecasting instructions follow. The MM output clause asks for the complete correct set:
\begin{apppanel}{MM output clause (verbatim)}
\raggedright
Your final answer MUST end with this exact format: listing all correct option(s) you have identified, separated by commas, within the box. For example: \texttt{\textbackslash boxed\{A\}} for a single correct option, or \texttt{\textbackslash boxed\{B, C\}} for multiple correct options.
\end{apppanel}
Partial F1 credit (\cref{app:method-eval}) is applied only offline; the prompt never asks the model to optimize a partial-credit surrogate. Reproduction should use the stored rendered prompts, because renderer fragments and runtime instructions are separate parts of the run contract.

The scaffold asks for decomposition, several distinct search angles, reflection on each result, corroboration across sources, a counter-case, and a final commitment. These are instructions, not evidence that every trace follows them: the Codex Default trial in \cref{app:case-monaco} answers without searching.

The model controls only the query string. The function interface is pinned to \texttt{web\_search} with one required string field, \texttt{query}, and \texttt{additionalProperties} set to \texttt{false}. The harness computes $\chi_i=\tau_i-1$ day and passes it to the retrieval client as \texttt{end\_date}; no date field appears in the model-visible schema. Send-time checks reject additional tool families, provider plugin fields, and online-browsing routes, and the filter model is called without any search tool. These checks constrain the exposed interfaces but cannot rule out undisclosed provider-side retrieval (\cref{app:residual-channels}).

\subsection{Budget and Termination Semantics}
\label{app:protocol}

The step and search limits constrain different resources: a step may request several searches, but at most four searches are executed per trial. Once the search budget is spent, the harness removes the tool and requests a final answer; near the step limit it issues a warning, and it disables tools on the final step. A turn without a parseable boxed answer can receive a continuation instruction while budget remains, and a parseable answer ends the trial.

Status messages report the current step and the number of searches used. Tool failures are returned within the assistant--tool message cycle rather than as a separate user turn. An empty result list after filtering is a valid protocol outcome, distinct from a request that failed. The retained traces distinguish early commitment, budget exhaustion, filtering losses, and failed requests.

\subsection{Leakage Filter}
\label{app:detector}

\paragraph{Configuration.}
Depending on the source run, the filter model is \texttt{qwen3.6-flash-\allowbreak 2026-04-16} or \texttt{qwen3.8-flash}, with a 512- or 2,048-token output cap. All runs share one filter prompt, whose fingerprint \texttt{29e22a4c5bccc2bb} is the first 16 hexadecimal characters of its SHA-256.

\paragraph{Decision rule.}
The prompt instructs the filter to drop a result that mentions a realized event, a scheduled event, or speculation tied to a time strictly after the cutoff, to drop when uncertain, and to judge the supplied text rather than its own knowledge. The filter receives only the cutoff and each result's title, URL, publication date, extracted content, and raw page content when available; it never sees the question, candidates, gold label, or forecasting dialogue. It replies with a JSON object holding a \texttt{keep} or \texttt{drop} verdict and a short reason, and a drop removes the whole result, including its title and URL.

\paragraph{Failures.}
The verdict parser accepts a valid object on its own or embedded in surrounding text. Invalid JSON and invalid verdicts are failures, not implicit keeps: transient failures are retried as configured, and a result still lacking a valid verdict is removed (fail-closed). When retrieval infrastructure is unhealthy, collection-health checks can instead mark the observation incomplete, which remains distinguishable from an empty result set produced because every candidate was judged unsafe.

\paragraph{Records.}
Each retrieval record links the query and cutoff to the raw candidate count, the retained count, the per-item verdicts in order, available reasons, and publication dates. Filter explanations stay in the record and never reach the forecasting model. When every result is removed, the implementation also clears the search provider's synthesized answer, so that the summary cannot bypass the empty result set. Raw payloads and reasons are not available for every source record, and a missing field cannot be reconstructed from a retained count alone.

\subsection{Configuration Provenance}
\label{app:configuration-provenance}

Trials for the 27 configurations come from 45 source runs, and provider-side model updates between runs cannot be detected from the request records. The configured admission dates are operational metadata, not a reconstruction of provider training corpora. Some source runs omit the field, and a missing value is not evidence that a model lacks prior knowledge. The records also distinguish an unspecified provider default, an explicit effort request, and a route whose name contains an effort label, such as GPT-5.4 High.

These distinctions limit the reasoning comparison. The Qwen and Kimi On settings are activation switches rather than a common High compute budget, and the two GLM-5 configurations are separate recorded requests. GPT-5.4 Default uses temperature 0.7 and its High route 1.0, so that contrast does not isolate reasoning effort from sampling. Zero reported reasoning tokens are not relabelled as an explicit Off request.

\section{Scoring and Statistical Definitions}
\label{app:method-eval}

\paragraph{Parsing and validity.}
The parser reads the last \texttt{\textbackslash boxed\{...\}} payload. Yes/no strings are matched case-insensitively to their fixed candidate labels, and named-binary strings must match one of the two stored outcome names. Multiple-choice payloads are split on commas or whitespace, checked against the candidate letters, and deduplicated into a set. An empty payload, an unknown label, or a missing box makes the output invalid ($\bot$). Refusals and invalid outputs score zero and remain in the denominator, whereas missing slots and unresolved infrastructure failures make a configuration incomplete, and an incomplete configuration receives no official Score. In the reported panel, all 27 configurations have three protocol outcomes for each of the 300 questions.

\paragraph{Item credit.}
For gold set $G$ and predicted set $\widehat G$, $\mathrm{TP}=|G\cap\widehat G|$, $\mathrm{FP}=|\widehat G\setminus G|$, and $\mathrm{FN}=|G\setminus\widehat G|$. Binary and MCS credit is exact set equality, so selecting several letters on a single-answer item scores zero. MM credit is the set F1, $2\mathrm{TP}/(2\mathrm{TP}+\mathrm{FP}+\mathrm{FN})$: for $G=\{A,C,D,H\}$, predicting $\{D\}$ gives $\mathrm{TP}=1$, $\mathrm{FP}=0$, $\mathrm{FN}=3$, credit 0.40, and 1.20 of three points, while predicting $\{A,C,D,E,J,K\}$ gives $\mathrm{TP}=3$, $\mathrm{FP}=3$, $\mathrm{FN}=1$, credit 0.60, and 1.80 points. Neither prediction is an exact match.

\paragraph{Aggregate scores.}
Score averages the three trial credits within each question, weights questions by one, two, or three points, and divides by the 528-point total. Exact@1 averages exact correctness over trials and questions. The prompt asks for the exact correct set, whereas offline MM scoring also credits partially correct sets. The analytic random-guess baseline picks one option uniformly on binary and MCS items and selects each MM option independently with probability $1/2$; its expected Score is 34.12.

\begin{table*}[t]
 \centering\footnotesize
 \setlength{\tabcolsep}{4pt}
 \appendixdatatable{\input{experiment_assets/tables/complete_results.tex}}
 \caption{All 27 configurations on 300 questions, in Score order. 95\% CI: paired bootstrap (\cref{app:significance}). Exact@1: strict accuracy (\%). Vote Score: strict complete-set majority. Cost: USD per 300-question trial at fixed public tariffs (\cref{app:cost}). CT: custom tools.}
 \label{tab:complete-results}
\end{table*}

\begin{table*}[t]
 \centering\footnotesize
 \setlength{\tabcolsep}{4pt}
 \appendixdatatable{\input{experiment_assets/tables/type_results.tex}}
 \caption{Exact accuracy by question type and MM F1 credit (\%), in Score order. Binary pools yes/no and named-binary questions.}
 \label{tab:type-results}
\end{table*}

\begin{table*}[t]
 \centering\footnotesize
 \setlength{\tabcolsep}{3.5pt}
 \appendixdatatable{\input{experiment_assets/tables/selection_errors.tex}}
 \caption{MM diagnostics in Score order: exact accuracy, F1 credit, omission rate $e_{\mathrm{om}}$, and false-selection rate $e_{\mathrm{fs}}$ (\%), with mean missed (FN) and extra (FP) options per trial (\cref{app:mm-diagnostics}).}
 \label{tab:selection-errors}
\end{table*}

\paragraph{Repeated-trial metrics.}
Vote Score applies strict voting to complete normalized answer sets: with three trials, a set needs at least two identical votes, invalid outputs abstain, and a question without a majority scores zero. Voting on each option separately could instead assemble a set that no trial proposed. Pass@3 and All@3 record whether at least one or all three trials are exact. Oracle Score takes the highest-credit trial of each question and thus requires the gold label. Vote coverage is the share of questions with a complete-set majority, agreement is the mean share of identical answer sets over the three pairs of trials, and Same wrong is the share of questions with three identical incorrect sets.

\paragraph{Evaluation chain.}
The four levels of $\mathcal{E}_M$ (\cref{sec:method}) map to the reported quantities as follows. \emph{Validity}: the parser returns a letter set or $\bot$. \emph{Item}: each valid set yields TP, FP, FN, exact correctness, and credit. \emph{Question}: the three trials of a question yield mean credit, Pass@3, All@3, the strict vote, agreement, the same-wrong indicator, and the best trial's credit. \emph{Model}: weighted aggregation over the 300 questions yields Score, Exact@1, Vote Score, Oracle Score, accuracy by question type, MM error rates (\cref{app:mm-diagnostics}), and cost (\cref{app:cost}).

\subsection{Multiple-Answer Error Severity}
\label{app:mm-diagnostics}

For an MM item with $g=|G|$ correct options among $m=|\mathcal{A}|$ candidates, the omission and false-selection rates are
\begin{equation}
 e_{\mathrm{om}}=\frac{\mathrm{FN}}{g},\qquad
 e_{\mathrm{fs}}=\frac{\mathrm{FP}}{m-g}.
 \label{eq:mm-error-severity}
\end{equation}
One omission among four correct options gives $e_{\mathrm{om}}=0.25$, and two give $0.50$; selecting one of six incorrect options gives $e_{\mathrm{fs}}=1/6$, and two give $2/6$. The rates thus capture the extent of an error, not only its occurrence. MM recall is $100(1-e_{\mathrm{om}})$ and incorrect-option rejection (specificity) is $100(1-e_{\mathrm{fs}})$; their radar scaling is given in \cref{app:profile-selection}.

We also report the mean numbers of missed and extra options, FN and FP. Rates and counts are averaged over valid MM trials within each question, then over questions with at least one valid trial. MM exact accuracy and F1 credit use all three trials, with invalid responses scored zero.

\subsection{Confidence Intervals}
\label{app:significance}

Intervals use 2,000 paired question-cluster bootstrap replicates, stratified by the four question types in \cref{tab:question-distribution}, with seed 20261001. Each resampled question carries all three of its trials, and all configurations share the same resampled indices, so differences are paired. The 2.5th and 97.5th percentiles give Score and paired-difference intervals (\cref{tab:complete-results,tab:paired-deltas}), which are not adjusted for multiple comparisons. The boundary comparison (\cref{tab:boundary-comparison}) uses the same procedure.

\section{Full Results and Multiple-Answer Diagnostics}
\label{app:type-analysis}

\Cref{tab:complete-results} reports Score, Exact@1, Vote Score, and cost for all 27 configurations; \cref{tab:type-results} gives accuracy by question type and MM F1 credit, with binary accuracy pooling the 109 yes/no and named-binary questions; and \cref{tab:selection-errors} reports the MM error-severity measures of \cref{app:mm-diagnostics}. All three tables list configurations in Score order.

\subsection{Representative Configurations and Radar Scaling}
\label{app:profile-selection}

The six profiles in \cref{fig:answer-profile} are the configurations that attain an observed extremum over all 27. GPT-5.4 High has the highest Score and binary accuracy; Claude Opus 4.6 the highest MCS accuracy and All@3; Claude Sonnet 4.6 the highest MM F1 credit, MM exact accuracy, and MM recall; and GPT-OSS 120B Default the highest incorrect-option rejection. GPT-OSS 120B Low has the lowest estimated forecasting cost, and Gemini 3.1 Flash-Lite Minimal the shortest mean duration. Tied extrema are retained. The figure script exports the selected values and criteria to \texttt{diagnostic\_profiles.csv}.

Each radar axis maps its panel minimum and maximum to radius zero and one: the plotted radius is the observed value minus the minimum, divided by the range. Because each axis has its own affine scale, polygon area is not an aggregate metric. \Cref{tab:profile-axis-ranges} gives the endpoints; raw values are in \cref{tab:type-results,tab:selection-errors,tab:repeat-results}.

\begingroup
\setlength{\intextsep}{6pt}
\begin{table}[!tbp]
 \centering\footnotesize
 \captionsetup{skip=4pt}
 \renewcommand{\arraystretch}{0.96}
 \appendixdatatable{\input{experiment_assets/tables/profile_axis_ranges.tex}}
 \caption{Radar scaling endpoints over all 27 configurations (percentage points). Binary and MCS denote exact accuracy; All@3 requires all three trials to be exact.}
 \label{tab:profile-axis-ranges}
\end{table}
\endgroup

The cost panel places all 27 configurations on a logarithmic dollar axis, as does \cref{fig:score-cost}. A configuration lies on the frontier if no other configuration has at least as high a Score at no greater cost, with at least one strict improvement. Frontier membership uses point estimates; the paired intervals in \cref{tab:paired-deltas} quantify the uncertainty of Score comparisons. Mean durations include model interaction and tool waits under the recorded serving conditions.

\section{Repeated Trials}
\label{app:repetition}

\Cref{tab:repeat-results} reports the repeated-trial metrics defined in \cref{app:method-eval}. Repetition separates what a configuration can reach from what it reaches reliably: Pass@3 exceeds Exact@1 by 12.4 to 20.2 percentage points, All@3 falls 13.1 to 21.0 points below it, and Oracle Score exceeds Score by 11.7 to 20.7 points. Strict Vote Score is lower than Score for all 27 configurations (\cref{tab:complete-results}), by 1.1 (GPT-OSS 120B Default) to 7.8 points (Qwen 3.5 Plus On): partly correct MM sets often lack a majority, and a stable incorrect set survives voting. Agreement is likewise no guarantee of correctness, as 10.3\% to 19.7\% of questions receive the same incorrect set in all three trials.

\begin{table*}[t]
 \centering\footnotesize
 \setlength{\tabcolsep}{3.5pt}
 \appendixdatatable{\input{experiment_assets/tables/repeat_results.tex}}
 \caption{Repeated-trial diagnostics (\%), in Score order. Oracle Score uses the highest-credit trial; Vote Score is given in \cref{tab:complete-results}.}
 \label{tab:repeat-results}
\end{table*}

\section{Representative Process Cases}
\label{app:cases}

The three cases link answer-set errors to the recorded queries, retained evidence, and final answers. Each panel lists the inspected trials, with Kept giving the results retained by each search out of five candidates; every trial is identified by question ID, configuration, and trial index in the case records (\cref{app:persistence}).

\subsection{Hugo Nominations: Omission and Addition}
\label{app:case-hugo}

\begin{apppanel}{2026 Hugo Award for Best Series nominees (MM)}
\casemeta{April 16, 2026}{April 15, 2026}{$\{A,C,D,H\}$: Cemeteries of Amalo, Emily Wilde, October Daye, Old Man's War.}
\begin{casetrials}
GPT-OSS 120B Default, trial 2 & 1, 2, 0, 1 & $\{D\}$ & 0.40\\
Claude Sonnet 4.6, trial 1 & 1, 2, 0, 0 & $\{A,C,D,E,J,K\}$ & 0.60\\
\end{casetrials}
\vspace{7pt}
{\footnotesize
\begin{tabularx}{\linewidth}{@{}rLr@{}}
\toprule
\# & Recorded query & Kept\\
\midrule
\multicolumn{3}{@{}l}{\textit{GPT-OSS 120B Default, trial 2}}\\
1 & 2026 Hugo Award Best Series nominees & 1\\
2 & 2026 Hugo Awards Best Series nominees & 2\\
3 & 2026 Hugo Award Best Series nominees list & 0\\
4 & ``Best Series'' Hugo 2026 nominees & 1\\
\midrule
\multicolumn{3}{@{}l}{\textit{Claude Sonnet 4.6, trial 1}}\\
1 & Hugo Award Best Series 2026 nominations predictions eligible & 1\\
2 & Hugo Award Best Series eligibility rules 2025 2026 & 2\\
3 & 2026 Hugo Award finalists nominees Best Series LACon V announced & 0\\
4 & site:lacon.org Hugo 2026 finalists Best Series & 0\\
\bottomrule
\end{tabularx}}
\end{apppanel}

GPT-OSS 120B searches for the nominee list four times, but the retained pages mainly cover nomination procedures and eligibility, and one call retains nothing. Its final answer selects only October Daye ($\mathrm{TP}=1$, $\mathrm{FN}=3$), and its three trials return $\{D,I\}$, $\{D\}$, and $\{D\}$: repeated searching adds little nominee evidence. The trace also records a filter error, in which a page from September 2025 is judged later than the April 2026 cutoff and removed, showing how a filtering mistake can shrink the evidence available for answer selection.

Claude Sonnet 4.6 searches eligibility rules and finalist announcements but also receives no verified nominee list. Its final explanation combines eligibility rules with popularity, author reputation, and recent activity; it dismisses Old Man's War as an older series and adds Singing Hills, Midsolar Murders, and Martin Hench. All three trials return the same six-letter set, whose three correct choices earn partial credit while three additions and one omission preclude an exact match.

\subsection{Champions League: Correct Size, Wrong Member}
\label{app:case-ucl}

\begin{apppanel}{2025--26 UEFA Champions League finalists (MM)}
\casemeta{May 7, 2026}{May 6, 2026}{$\{C,G\}$: PSG, Arsenal.}
\begin{casetrials}
Qwen 3.5 Plus On, trial 1 & 2, 2, 4, 2 & $\{E,G\}$ & 0.50\\
\end{casetrials}
\end{apppanel}

Qwen 3.5 Plus On predicts Bayern and Arsenal ($\mathrm{TP}=\mathrm{FP}=\mathrm{FN}=1$), earning 1.50 of three points. The set has the correct size, yet both recall and rejection register the substitution. The final explanation relies on forecasts and betting assessments, including an Opta forecast favoring Arsenal and Bayern; such pre-outcome forecasts are permissible evidence. The three trials return $\{E,G\}$, $\{C,H\}$, and $\{E,G\}$, so the imperfect membership reflects uncertainty in the evidence used.

\subsection{Monaco: Evidence and the Cutoff Boundary}
\label{app:case-monaco}

\begin{apppanel}{2026 Monaco Grand Prix top three (MM)}
\casemeta{June 8, 2026}{June 7, 2026}{$\{A,E,I\}$.}
\begin{casetrials}
GPT-5.3 Codex Default, trial 1 & no search & $\{C,D,F\}$ & 0.00\\
GPT-5.3 Codex Low, trial 1 & 4, 3 & $\{A,E,I\}$ & 1.00\\
\end{casetrials}
\end{apppanel}

GPT-5.3 Codex Default answers without searching, predicting $\{C,D,F\}$ from general driver and team strength and the importance of qualifying; all three Default trials return this set and earn zero credit. GPT-5.3 Codex Low makes two searches, the second asking for the Monaco results; it retains a June 7 report naming the top three drivers, and all three Low trials are exact. The outcome was thus reported on the permitted cutoff day (\cref{app:method-replay}). The case explains this retrieval difference and shows why the first public disclosure of an answer must be checked alongside its recorded resolution date.

\section{Costs and Resources}
\label{app:cost}

\begin{table*}[!tp]
 \centering
 \begin{minipage}[t]{0.49\textwidth}
 \vspace{0pt}\centering\footnotesize
 \setlength{\tabcolsep}{2pt}
 \renewcommand{\arraystretch}{1.0}
 \appendixheader{Model / billing service & \shortstack[l]{Input\\tier} & \shortstack[r]{Input\\/ M} & \shortstack[r]{Output\\/ M}}
 \begingroup
 \renewenvironment{tabular}[1]{%
  \begin{tabular*}{\linewidth}{@{\extracolsep{\fill}}>{\raggedright\arraybackslash}p{0.42\linewidth}lrr@{}}%
 }{\end{tabular*}}%
 \input{experiment_assets/tables/unit_prices.tex}
 \endgroup
 \caption{Forecasting-model tariffs in USD per million tokens, by per-request input-length tier, as of October 6, 2026.}
 \label{tab:unit-prices}
 \end{minipage}\hfill
 \begin{minipage}[t]{0.49\textwidth}
 \vspace{0pt}\centering\footnotesize
 \setlength{\tabcolsep}{2pt}
 \appendixheader{Configuration & \shortstack[r]{Three\\trials (\$)} & \shortstack[r]{Per trial\\(\$)}}
 \appendixdatatable{\input{experiment_assets/tables/total_costs.tex}}
 \caption{Forecasting-model cost on 300 questions: the recorded three-trial total and the expected cost per trial (\cref{eq:normalized-cost}), in Score order.}
 \label{tab:total-costs}
 \end{minipage}
\end{table*}

\begin{figure}[!tbp]
 \centering
 \includegraphics[width=\columnwidth]{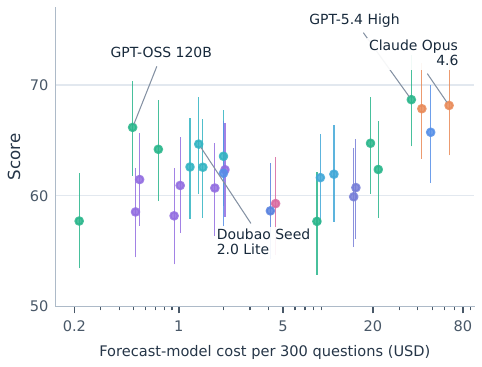}
 \caption{Score against forecasting-model cost at public tariffs for all 27 configurations; the cost axis is logarithmic.}
 \label{fig:score-cost}
\end{figure}

\paragraph{Forecasting-model cost.}
We reprice every recorded forecasting call at fixed standard online public tariffs (\cref{tab:unit-prices}), applying input-length tiers per request. For call $j$ with $t_j^{\mathrm{in}}$ input and $t_j^{\mathrm{out}}$ output tokens and per-million prices $p_j^{\mathrm{in}}$ and $p_j^{\mathrm{out}}$, the expected cost of one 300-question trial is
\begin{equation}
 \mathrm{Cost}_{300}=\frac{1}{3}\sum_j
 \frac{p_j^{\mathrm{in}}t_j^{\mathrm{in}}+p_j^{\mathrm{out}}t_j^{\mathrm{out}}}{10^6},
 \label{eq:normalized-cost}
\end{equation}
where $j$ ranges over a configuration's calls in all three trials. Output tokens include reported reasoning tokens. Prices are frozen as of October 6, 2026, with exact-model historical tariffs where required; GPT-OSS 120B uses the DeepInfra hosted tariff and Kimi K2.5 AliCloud pricing, and yuan prices are converted at 6.7048281 CNY per USD. Source URLs and native prices are kept in the pricing data. Search and filter charges are excluded. \Cref{tab:total-costs} reports the results, and \cref{fig:score-cost} plots them against Score. Across all 27 configurations, the 96,299 recorded forecasting calls cost \$949.50 for the three trials.

\paragraph{Stage accounting.}
The request ledger records 491,217 requests: 84,188 forecasting, 333,009 leakage-filter, 415 diagnostic-probe, and 73,605 search requests. \Cref{tab:stage-costs} reprices the returned token usage by role and counts requests returned without usage. The ledger covers one accounting period rather than every recorded forecasting call, so its repriced forecasting total (\$774.97) is lower than the \$949.50 above. Search charges require separate billing records, and eight probe requests and six filter requests with usage lack a supported public tariff. Reported invoices total \$830.80, including \$104.09 for the filter. The recorded Codex invoice of \$8.70 comprises \$3.31 in the request ledger and \$5.39 pending attribution.

\paragraph{Resources.}
\Cref{tab:resources} gives the mean searches, model steps, tokens, and latency per prediction. Latency is the stored run duration, including model interaction and tool waits under the recorded serving conditions.

\begin{table*}[!tbp]
 \centering\small
 \appendixdatatable{\input{experiment_assets/tables/stage_costs.tex}}
 \caption{Ledger requests and repriced token usage by stage. Missing token usage counts requests returned without usage; search is billed separately.}
 \label{tab:stage-costs}

 \vspace{10pt}
 \centering\footnotesize
 \setlength{\tabcolsep}{3.5pt}
 \appendixdatatable{\input{experiment_assets/tables/resource_results.tex}}
 \caption{Mean recorded resources per prediction, in Score order. Tokens are in thousands, and output includes reported reasoning tokens. Latency is the stored run duration.}
 \label{tab:resources}
\end{table*}

\section{Protocol Positioning and Information Boundaries}
\label{app:leakage}

The three analyses below expand the boundary evidence in \cref{sec:experiment}: the criteria behind \cref{tab:positioning}, the controlled comparison in \cref{tab:boundary-comparison}, and the residual-leakage audit.

\subsection{Positioning Criteria}
\label{app:positioning}

\Cref{tab:positioning} grades each design with one rule per column, applied identically to every row, ours included. A property earns credit only when the cited paper or dataset documentation states it, and an audit whose labellers are not identified does not count as human-verified. Cutoff and Leak do not apply (--) to live designs, whose outcomes are unknown to every model.

\begin{apppanel}[breakable]{Grading rules for \cref{tab:positioning}}
\raggedright
\newcommand{\rubric}[4]{\textbf{#1.}\enspace\cmark{}~#2\enspace\pmark{}~#3\enspace\xmark{}~#4\par\smallskip}
\rubric{Live}{Forecasts are made before the outcomes are known.}{The design's own evaluation adds live forecasts to a mainly retrospective one.}{The design evaluates resolved questions only.}
\rubric{Reuse}{A resolved question stays valid for newer models without new questions or evidence.}{Resolved questions are replayed under a temporal boundary, but newer models need new questions or crawls, or their validity is not ensured.}{Resolved questions are dropped, or reused without a temporal boundary.}
\rubric{Web}{Agents query a general-purpose live web search at inference time.}{Live retrieval is limited to news APIs or to pre-collected pages.}{Live web retrieval is not part of the design.}
\rubric{Snap.}{Evidence is frozen in a corpus or snapshot that any query can be replayed against.}{Per-trial queries and payloads are archived and replayable, without a frozen corpus.}{Retrieved evidence is not archived for replay.}
\rubric{Cutoff}{Questions are admitted per model and question from the model's knowledge cutoff.}{Parametric knowledge is bounded by one test window or by model selection.}{Parametric knowledge is not bounded.}
\textbf{Leak.}\enspace\cmark{}~Retrieved content is filtered before the model reads it, blind to the answer, and audited with human-verified labels.\enspace\pmark{}~A content-level check exists but is LLM-only, runs after forecasting, or conditions on the gold answer.\enspace\xmark{}~No content-level check; date restrictions alone do not count.
\end{apppanel}

\paragraph{Borderline cells.}
Web separates live search engines from fixed sources: BTF's retrieval issues live search-engine queries but returns only pre-scraped pages~\citep{wildman2025btf}, whereas FutureSim's agent has no network access and searches an offline news snapshot~\citep{goel2026futuresim}. ForecastBench and Halawi et al. retrieve through news APIs~\citep{karger2025forecastbench,halawi2024approaching}. Leak counts content-level checks only: Halawi et al., BTF, and FutureSim gate evidence by publication or crawl date, which acts on metadata rather than content. Among the partial Leak cells, BLF filters content before the model reads it but relies on LLM verdicts throughout~\citep{murphy2026agentic}; AIA Forecaster's judge reads the finished forecast together with the resolution, and flagged responses are removed afterwards~\citep{alur2025aia}; and Agentic Time Machine conditions its filter on the resolved answer, without identifying who labelled its audit sample~\citep{chai2026timemachine}. The same paper never screens models by knowledge cutoff, so its Cutoff cell is \xmark{}. AIA Forecaster's Live cell is partial because one week of live forecasts supplements its backtests. FutureX-Past releases question--answer pairs without a retrieval setting and warns that newer models may know the outcomes~\citep{futurexpast2025}. OracleProto's Snap.\ cell is partial: it archives the queries and payloads of every trial, which \cref{app:boundary-comparison} replays, but no corpus answers arbitrary new queries.

\subsection{Controlled Boundary Comparison}
\label{app:boundary-comparison}

\paragraph{Models and conditions.}
GLM-5, Kimi K2.5, Doubao Seed 2.0 Lite, and Qwen 3.5 Flash are queried through OpenRouter with default provider routing; \cref{tab:boundary-banks} lists the route identifiers. Each model answers the 300 questions three times under each of the three conditions, with prompts, sampling settings, and budgets fixed across conditions. Because the routes and the retrieval mechanism differ from the main benchmark, these Scores are not interchangeable with those in \cref{tab:complete-results}.

\paragraph{Frozen candidate banks.}
Both retrieval conditions draw from one frozen bank per model, question, and trial. A bank holds the unfiltered Tavily results of the searches recorded in the corresponding main-benchmark trial, under the original cutoff; where these raw results had not been retained, the recorded queries were reissued once with the original cutoff and search settings, and the responses were frozen. A query identical to a recorded one returns its saved batch, and any other query returns the five bank results with the highest length-normalized lexical overlap, with ties broken deterministically. Banks never share results across questions, trials, models, or cutoffs.

\paragraph{Filtering.}
Under $D$, the selected results are shown unfiltered. Under $F$, Qwen 3.8 Flash verdicts are applied after selection, and dropped results are not replaced. Verdicts saved with the source trial are reused; missing verdicts were obtained with the same filter prompt at temperature 0, with a 2,048-token cap, reasoning disabled, and JSON output, through OpenRouter or, for the remaining items, Alibaba Cloud's official API. Candidates that the filter provider refused to judge under its content policy are dropped (\cref{tab:boundary-banks}).

\begin{table}[!tbp]
 \centering\footnotesize
 \setlength{\tabcolsep}{3pt}
 \newcommand{\route}[1]{\multicolumn{6}{@{}l}{\scriptsize\texttt{#1}}\\[2pt]}
 \begin{tabular*}{\columnwidth}{@{\extracolsep{\fill}}lrrrrr@{}}
 \toprule
 Model & Calls & Raw & Kept & \% & Refused\\
 \midrule
 GLM-5 & 3,559 & 15,957 & 6,828 & 42.8 & 7\\
 \route{z-ai/glm-5}
 Kimi K2.5 & 3,541 & 15,829 & 6,418 & 40.5 & 2\\
 \route{moonshotai/kimi-k2.5}
 Doubao Seed 2.0 Lite & 3,041 & 13,845 & 5,588 & 40.4 & 4\\
 \route{bytedance-seed/seed-2.0-lite}
 Qwen 3.5 Flash & 3,478 & 15,650 & 6,671 & 42.6 & 2\\
 \route{qwen/qwen3.5-flash-02-23}
 \bottomrule
 \end{tabular*}
 \caption{Frozen candidate banks, 900 per model (300 questions, three trials), with each model's OpenRouter route. Calls: recorded searches; Raw: candidate result occurrences; Kept: occurrences the filter keeps, counted before query selection; \%: Kept/Raw, the evidence retained in \cref{tab:boundary-comparison}; Refused: candidates dropped after a content-policy refusal.}
 \label{tab:boundary-banks}
\end{table}

\paragraph{Interpretation.}
The bank fixes the evidence pool but not the trajectory: under $D$ and $F$, a model can issue different queries after reading different evidence, and lexical ranking stands in for the search engine on unseen queries. The contrasts therefore measure the replay of recorded evidence under each boundary, not the effect of fresh searches. Retention counts evidence occurrences and estimates neither useful information nor leakage.

\subsection{Residual Leakage Audit}
\label{app:audit}

\paragraph{Criterion.}
A retained search result counts as leakage when it states a realized fact dated after the inclusive cutoff; schedules, predictions, and speculation about later events are permitted. Reviewers judge the event time of the stated fact from the page text and supporting dates. This criterion is looser than the filter's rule, which also drops scheduled events and speculation after the cutoff (\cref{app:detector}). The audit therefore measures leakage among retained results, not how much permissible evidence the filter removed.

\paragraph{Sample and results.}
We randomly sample 10,000 post-filter results from all main-benchmark trials (seed 20262026). GPT 6.1 Sol labels 137 of them as leakage (\cref{tab:leakage-audit}). Human reviewers confirm all 137 positives and find no leakage in 1,000 sampled negatives, so we adopt the resulting rate of 1.37\%.

\begin{table}[!tbp]
 \centering\small
 \begin{tabular}{@{}lrr@{}}
 \toprule
 Review & Checked & Confirmed leaks\\
 \midrule
 GPT 6.1 Sol & 10,000 & 137\\
 Human: labelled positive & 137 & 137\\
 Human: labelled negative & 1,000 & 0\\
 \bottomrule
 \end{tabular}
 \caption{Residual-leakage audit of a random sample of post-filter results, with human verification. The adopted leakage rate is 1.37\%.}
 \label{tab:leakage-audit}
\end{table}

\paragraph{Interpretation.}
The human check is stratified by the auditor's label: all 137 positives and 1,000 sampled negatives. It supports the reported positives and the checked negative sample, but it is not a human adjudication of all 10,000 items and establishes neither population recall nor an exact zero false-negative rate. The Hugo date-comparison error (\cref{app:case-hugo}) and the Monaco disclosure boundary (\cref{app:case-monaco}) show why individual traces remain necessary alongside an aggregate rate.

\section{Reproducibility and Remaining Information Channels}
\label{app:method-channels}

The data manifest records source hashes, scoring definitions, and the data behind every figure and table. Archived configurations record admission dates, sampling settings, filter versions, and tool boundaries. Query traces and returned payloads support checks for outcome cues, first answer disclosure, provider updates, and changed pages. Scripts regenerate all figures and tables from these records.

\subsection{Artifacts Needed for Reproduction}
\label{app:persistence}

The manifest fingerprints the source database and source observations, lists hashes of the analysis inputs, and records the scoring contract. A reproduction must preserve the question IDs, option mapping, trial membership, item weights, treatment of invalid and missing outcomes, and bootstrap seed. Matching hashes identify the supplied bytes; they do not certify that a web page or deployed model can be queried again in the same state. Three record groups serve different checks:
\begin{description}[leftmargin=0pt,labelindent=0pt,labelsep=0.4em,itemsep=3pt,topsep=3pt]
\raggedright
\item[Configuration records.] Per-run admission dates, provider profiles, requested sampling fields, filter model and output cap, and prompt fingerprint establish the requested conditions.
\item[Trial and case records.] Question ID, configuration, trial index, predicted letters, source ID, trace hash, and search sequence tie each displayed case to one observed trial; raw responses and retained payloads support interpretation where available.
\item[Analysis records.] Per-configuration summaries, scores by question type, MM selection errors, paired comparisons, tariff tables, and request ledgers supply the inputs to the figures and tables, with separate aggregation rules for cost and quality.
\end{description}

Recomputing metrics from saved predictions differs from rerunning models against a live search backend: provider updates, changing search indexes, revised pages, and source-specific filter conditions can change a rerun even under identical budgets. The case records identify the inspected trials; their trajectories are not presented as typical model behavior.

\subsection{Remaining Information Channels}
\label{app:residual-channels}

\paragraph{Parametric knowledge and provider interfaces.}
A disclosed or configured date cannot verify what deployed model parameters contain, and request-level browsing checks cannot expose hidden caches or retrieval middleware. A stronger exclusion claim would need provider-side evidence beyond the archived request.

\paragraph{Publication metadata and changed pages.}
A page may lack a date, keep an old header after an update, or republish newer facts. Text-level filtering detects explicit post-cutoff events but can miss claims whose timing is implicit, so query payloads, source text, and item-level verdicts matter independently of the publication-date field.

\paragraph{Task-side disclosure.}
The question stem, candidate order, or source resolution convention can reveal information that no retrieval filter removes. A report published on the inclusive cutoff day can be correctly retained even though the question no longer represents an unresolved forecast. We report the fixed population and the Monaco case explicitly rather than hiding this limitation through post-hoc item removal.

%% file: experiment_assets/tables/complete_results.tex
\begin{tabular}{@{}llrrrrr@{}}
\toprule
Model & Setting & Score & 95\% CI & Exact@1 & Vote Score & Cost (\$) \\
\midrule
GPT-5.4 & High & 68.7 & [64.4, 72.7] & 65.3 & 64.0 & 36.17 \\
Claude Opus 4.6 & Default & 68.1 & [63.6, 72.3] & 63.4 & 65.6 & 64.74 \\
Claude Sonnet 4.6 & Default & 67.9 & [63.3, 72.0] & 63.2 & 65.4 & 42.53 \\
GPT-OSS 120B & Default & 66.2 & [61.8, 70.4] & 61.6 & 65.1 & 0.49 \\
Gemini 3.1 Pro (CT) & High & 65.7 & [61.2, 70.0] & 63.6 & 64.6 & 48.75 \\
GPT-5.3 Codex & Low & 64.7 & [60.1, 68.9] & 60.2 & 58.2 & 19.26 \\
Doubao Seed 2.0 Lite & Default & 64.7 & [60.2, 68.9] & 59.0 & 61.8 & 1.36 \\
GPT-OSS 120B & High & 64.2 & [59.5, 68.6] & 61.2 & 59.2 & 0.73 \\
Doubao Seed 2.0 Lite & High & 63.6 & [59.0, 67.8] & 58.9 & 57.6 & 1.99 \\
Doubao Seed 2.0 Lite & Low & 62.6 & [57.9, 67.0] & 57.1 & 57.7 & 1.19 \\
Doubao Seed 2.0 Lite & Off & 62.6 & [58.0, 66.9] & 56.9 & 57.3 & 1.45 \\
GPT-5.4 & Default & 62.4 & [58.0, 66.7] & 58.9 & 58.0 & 21.72 \\
Qwen 3.5 Plus & Off & 62.3 & [58.1, 66.6] & 57.3 & 57.5 & 2.04 \\
Gemini 3.1 Flash-Lite & Minimal & 62.0 & [57.2, 66.8] & 57.3 & 57.7 & 1.99 \\
Kimi K2.5 & On & 61.9 & [57.6, 66.3] & 58.2 & 57.0 & 10.96 \\
Kimi K2.5 & Off & 61.6 & [57.4, 65.6] & 57.8 & 54.9 & 8.89 \\
Qwen 3.5 Flash & On & 61.4 & [57.3, 65.7] & 58.0 & 56.1 & 0.55 \\
Qwen 3.5 35B A3B & On & 60.9 & [56.6, 65.2] & 56.0 & 56.1 & 1.02 \\
GLM-5 & On & 60.7 & [56.1, 65.1] & 55.3 & 53.0 & 15.38 \\
Qwen 3.5 Plus & On & 60.7 & [56.4, 64.8] & 56.4 & 52.9 & 1.75 \\
GLM-5 & Default & 59.9 & [55.3, 64.3] & 55.9 & 54.0 & 14.87 \\
MiniMax M2.5 & Default & 59.3 & [54.7, 63.5] & 56.9 & 55.0 & 4.46 \\
DeepSeek V3.2 Exp & Default & 58.6 & [54.3, 62.9] & 53.7 & 53.8 & 4.11 \\
Qwen 3.5 Flash & Off & 58.5 & [54.4, 62.5] & 55.0 & 52.8 & 0.51 \\
Qwen 3.5 35B A3B & Off & 58.2 & [53.8, 62.5] & 54.6 & 53.8 & 0.93 \\
GPT-OSS 120B & Low & 57.7 & [53.4, 62.0] & 54.7 & 53.6 & 0.22 \\
GPT-5.3 Codex & Default & 57.7 & [52.9, 62.2] & 53.3 & 52.7 & 8.43 \\
\bottomrule
\end{tabular}

%% file: experiment_assets/tables/type_results.tex
\begin{tabular}{@{}lrrrr@{}}
\toprule
Configuration & Binary exact & MCS exact & MM exact & MM credit \\
\midrule
GPT-5.4 / High & 74.6 & 65.2 & 38.7 & 72.6 \\
Claude Opus 4.6 / Default & 69.7 & 65.4 & 36.9 & 74.3 \\
Claude Sonnet 4.6 / Default & 69.7 & 63.4 & 43.2 & 78.3 \\
GPT-OSS 120B / Default & 69.1 & 62.6 & 35.1 & 73.3 \\
Gemini 3.1 Pro (CT) / High & 73.7 & 61.7 & 41.4 & 69.1 \\
GPT-5.3 Codex / Low & 68.2 & 60.2 & 36.9 & 74.0 \\
Doubao Seed 2.0 Lite / Default & 63.0 & 63.0 & 30.6 & 70.9 \\
GPT-OSS 120B / High & 71.6 & 60.8 & 32.4 & 66.3 \\
Doubao Seed 2.0 Lite / High & 67.0 & 59.5 & 32.4 & 71.4 \\
Doubao Seed 2.0 Lite / Low & 63.0 & 59.3 & 30.6 & 71.3 \\
Doubao Seed 2.0 Lite / Off & 60.2 & 59.5 & 36.0 & 73.3 \\
GPT-5.4 / Default & 69.7 & 59.1 & 26.1 & 64.2 \\
Qwen 3.5 Plus / Off & 65.7 & 57.8 & 30.6 & 71.6 \\
Gemini 3.1 Flash-Lite / Minimal & 62.7 & 59.5 & 32.4 & 68.3 \\
Kimi K2.5 / On & 67.3 & 58.7 & 29.7 & 65.8 \\
Kimi K2.5 / Off & 68.2 & 58.2 & 25.2 & 64.6 \\
Qwen 3.5 Flash / On & 69.4 & 56.5 & 30.6 & 67.4 \\
Qwen 3.5 35B A3B / On & 62.1 & 58.9 & 26.1 & 65.4 \\
GLM-5 / On & 64.2 & 56.1 & 26.1 & 70.2 \\
Qwen 3.5 Plus / On & 69.1 & 54.8 & 26.1 & 68.8 \\
GLM-5 / Default & 65.7 & 56.7 & 23.4 & 63.0 \\
MiniMax M2.5 / Default & 66.1 & 58.2 & 24.3 & 55.5 \\
DeepSeek V3.2 Exp / Default & 60.9 & 55.8 & 23.4 & 64.1 \\
Qwen 3.5 Flash / Off & 66.4 & 54.3 & 24.3 & 62.5 \\
Qwen 3.5 35B A3B / Off & 65.1 & 54.3 & 24.3 & 62.0 \\
GPT-OSS 120B / Low & 67.9 & 52.4 & 25.2 & 62.4 \\
GPT-5.3 Codex / Default & 62.4 & 55.6 & 17.1 & 58.7 \\
\bottomrule
\end{tabular}

%% file: experiment_assets/tables/selection_errors.tex
\begin{tabular}{@{}lrrrrrr@{}}
\toprule
Configuration & MM exact & MM credit & Omission & False selection & Mean FN & Mean FP \\
\midrule
GPT-5.4 / High & 38.7 & 72.6 & 26.8 & 15.3 & 0.97 & 0.87 \\
Claude Opus 4.6 / Default & 36.9 & 74.3 & 23.8 & 20.2 & 0.96 & 0.95 \\
Claude Sonnet 4.6 / Default & 43.2 & 78.3 & 18.1 & 18.3 & 0.87 & 0.91 \\
GPT-OSS 120B / Default & 35.1 & 73.3 & 28.0 & 11.8 & 1.21 & 0.68 \\
Gemini 3.1 Pro (CT) / High & 41.4 & 69.1 & 31.4 & 13.3 & 1.27 & 0.71 \\
GPT-5.3 Codex / Low & 36.9 & 74.0 & 24.4 & 18.7 & 1.04 & 0.95 \\
Doubao Seed 2.0 Lite / Default & 30.6 & 70.9 & 24.4 & 21.4 & 0.89 & 1.08 \\
GPT-OSS 120B / High & 32.4 & 66.3 & 31.1 & 17.4 & 1.30 & 0.96 \\
Doubao Seed 2.0 Lite / High & 32.4 & 71.4 & 26.2 & 20.3 & 1.25 & 1.09 \\
Doubao Seed 2.0 Lite / Low & 30.6 & 71.3 & 25.1 & 22.8 & 1.20 & 1.12 \\
Doubao Seed 2.0 Lite / Off & 36.0 & 73.3 & 24.2 & 19.1 & 1.11 & 1.03 \\
GPT-5.4 / Default & 26.1 & 64.2 & 35.2 & 21.1 & 1.30 & 1.23 \\
Qwen 3.5 Plus / Off & 30.6 & 71.6 & 26.4 & 20.3 & 1.05 & 0.94 \\
Gemini 3.1 Flash-Lite / Minimal & 32.4 & 68.3 & 25.7 & 25.8 & 0.79 & 1.19 \\
Kimi K2.5 / On & 29.7 & 65.8 & 32.9 & 17.8 & 1.34 & 0.99 \\
Kimi K2.5 / Off & 25.2 & 64.6 & 30.3 & 26.5 & 1.22 & 1.26 \\
Qwen 3.5 Flash / On & 30.6 & 67.4 & 31.1 & 19.0 & 1.19 & 1.11 \\
Qwen 3.5 35B A3B / On & 26.1 & 65.4 & 33.9 & 19.3 & 1.33 & 1.04 \\
GLM-5 / On & 26.1 & 70.2 & 29.0 & 15.9 & 1.14 & 0.89 \\
Qwen 3.5 Plus / On & 26.1 & 68.8 & 32.3 & 17.2 & 1.23 & 0.89 \\
GLM-5 / Default & 23.4 & 63.0 & 38.1 & 17.9 & 1.51 & 1.01 \\
MiniMax M2.5 / Default & 24.3 & 55.5 & 41.3 & 17.3 & 1.57 & 0.98 \\
DeepSeek V3.2 Exp / Default & 23.4 & 64.1 & 32.4 & 21.7 & 1.27 & 1.09 \\
Qwen 3.5 Flash / Off & 24.3 & 62.5 & 36.7 & 21.4 & 1.37 & 1.11 \\
Qwen 3.5 35B A3B / Off & 24.3 & 62.0 & 38.1 & 20.6 & 1.46 & 1.06 \\
GPT-OSS 120B / Low & 25.2 & 62.4 & 35.1 & 21.2 & 1.46 & 1.16 \\
GPT-5.3 Codex / Default & 17.1 & 58.7 & 36.5 & 30.8 & 1.30 & 1.86 \\
\bottomrule
\end{tabular}

%% file: experiment_assets/tables/profile_axis_ranges.tex
\begin{tabular}{@{}lrr@{}}
\toprule
Metric & Minimum & Maximum\\
\midrule
Binary & 60.24 & 74.62\\
MCS & 52.38 & 65.37\\
MM F1 & 55.53 & 78.30\\
MM exact & 17.12 & 43.24\\
All@3 & 32.67 & 50.33\\
MM recall & 58.73 & 81.93\\
MM rejection & 69.22 & 88.20\\
\bottomrule
\end{tabular}

%% file: experiment_assets/tables/repeat_results.tex
\begin{tabular}{@{}lrrrrrr@{}}
\toprule
Configuration & Pass@3 & All@3 & Vote cov. & Agreement & Same wrong & Oracle Score \\
\midrule
GPT-5.4 / High & 79.7 & 49.7 & 93.3 & 72.2 & 12.0 & 82.2 \\
Claude Opus 4.6 / Default & 76.3 & 50.3 & 95.0 & 77.2 & 18.0 & 79.8 \\
Claude Sonnet 4.6 / Default & 75.7 & 50.0 & 94.7 & 76.0 & 16.7 & 79.6 \\
GPT-OSS 120B / Default & 77.3 & 42.7 & 91.7 & 66.7 & 10.7 & 80.4 \\
Gemini 3.1 Pro (CT) / High & 77.3 & 47.7 & 93.0 & 72.4 & 13.3 & 79.9 \\
GPT-5.3 Codex / Low & 75.3 & 46.7 & 92.7 & 72.4 & 15.7 & 78.4 \\
Doubao Seed 2.0 Lite / Default & 72.3 & 44.3 & 92.7 & 71.6 & 16.3 & 77.4 \\
GPT-OSS 120B / High & 76.0 & 46.0 & 91.3 & 71.0 & 14.0 & 77.9 \\
Doubao Seed 2.0 Lite / High & 73.3 & 45.3 & 90.3 & 72.1 & 17.7 & 77.5 \\
Doubao Seed 2.0 Lite / Low & 70.3 & 43.7 & 91.7 & 72.1 & 18.3 & 75.0 \\
Doubao Seed 2.0 Lite / Off & 72.7 & 42.3 & 90.7 & 70.9 & 18.3 & 77.9 \\
GPT-5.4 / Default & 76.3 & 40.7 & 91.0 & 66.1 & 13.0 & 79.4 \\
Qwen 3.5 Plus / Off & 76.0 & 38.7 & 89.3 & 65.6 & 14.7 & 80.1 \\
Gemini 3.1 Flash-Lite / Minimal & 72.3 & 43.0 & 92.3 & 71.3 & 16.7 & 76.0 \\
Kimi K2.5 / On & 75.0 & 42.0 & 92.0 & 68.9 & 15.0 & 78.5 \\
Kimi K2.5 / Off & 78.0 & 37.7 & 88.0 & 61.6 & 10.3 & 82.0 \\
Qwen 3.5 Flash / On & 76.0 & 40.0 & 88.0 & 64.0 & 12.0 & 78.8 \\
Qwen 3.5 35B A3B / On & 71.7 & 38.3 & 89.7 & 64.7 & 13.3 & 77.2 \\
GLM-5 / On & 70.0 & 41.3 & 87.3 & 68.4 & 17.7 & 75.7 \\
Qwen 3.5 Plus / On & 73.0 & 40.0 & 90.0 & 66.9 & 15.3 & 77.4 \\
GLM-5 / Default & 73.3 & 38.7 & 88.0 & 64.9 & 14.7 & 77.1 \\
MiniMax M2.5 / Default & 73.3 & 39.3 & 90.3 & 66.2 & 13.0 & 76.2 \\
DeepSeek V3.2 Exp / Default & 73.0 & 32.7 & 88.0 & 60.5 & 13.0 & 77.5 \\
Qwen 3.5 Flash / Off & 75.0 & 34.7 & 87.7 & 61.2 & 13.0 & 79.2 \\
Qwen 3.5 35B A3B / Off & 73.0 & 35.0 & 89.3 & 62.2 & 13.3 & 76.5 \\
GPT-OSS 120B / Low & 72.7 & 35.0 & 90.3 & 62.8 & 13.0 & 75.7 \\
GPT-5.3 Codex / Default & 68.3 & 38.3 & 91.7 & 69.2 & 19.7 & 72.9 \\
\bottomrule
\end{tabular}

%% file: experiment_assets/tables/unit_prices.tex
\begin{tabular}{@{}llrr@{}}
\toprule
Model / billing service & Input-length tier & Input / M & Output / M \\
\midrule
GPT-5.4 & $\leq 272k$ & 2.500 & 15.000 \\
GPT-5.4 & $>272k$ & 5.000 & 22.500 \\
GPT-5.3 Codex & all & 1.750 & 14.000 \\
Claude Sonnet 4.6 & all & 3.000 & 15.000 \\
Claude Opus 4.6 & all & 5.000 & 25.000 \\
Gemini 3.1 Pro & $\leq 200k$ & 2.000 & 12.000 \\
Gemini 3.1 Pro & $>200k$ & 4.000 & 18.000 \\
Gemini 3.1 Flash-Lite & all & 0.250 & 1.500 \\
Qwen 3.5 Flash & $\leq 128k$ & 0.029 & 0.287 \\
Qwen 3.5 Flash & $(128k,256k]$ & 0.115 & 1.147 \\
Qwen 3.5 Flash & $>256k$ & 0.172 & 1.720 \\
Qwen 3.5 Plus & $\leq 128k$ & 0.115 & 0.688 \\
Qwen 3.5 Plus & $(128k,256k]$ & 0.287 & 1.720 \\
Qwen 3.5 Plus & $>256k$ & 0.573 & 3.440 \\
Qwen 3.5 35B A3B & $\leq 128k$ & 0.057 & 0.459 \\
Qwen 3.5 35B A3B & $>128k$ & 0.229 & 1.835 \\
Kimi K2.5 (AliCloud) & all & 0.574 & 3.011 \\
GLM-5 & all & 1.000 & 3.200 \\
MiniMax M2.5 & all & 0.300 & 1.200 \\
Doubao Seed 2.0 Lite & $\leq 32k$ & 0.089 & 0.537 \\
Doubao Seed 2.0 Lite & $(32k,128k]$ & 0.134 & 0.805 \\
Doubao Seed 2.0 Lite & $>128k$ & 0.268 & 1.611 \\
DeepSeek V3.2 Exp & all & 0.298 & 0.447 \\
GPT-OSS 120B (DeepInfra) & all & 0.037 & 0.170 \\
\bottomrule
\end{tabular}

%% file: experiment_assets/tables/total_costs.tex
\begin{tabular}{@{}lrr@{}}
\toprule
Configuration & Three trials (\$) & Per trial (\$) \\
\midrule
GPT-5.4 / High & 108.50 & 36.17 \\
Claude Opus 4.6 / Default & 194.21 & 64.74 \\
Claude Sonnet 4.6 / Default & 127.60 & 42.53 \\
GPT-OSS 120B / Default & 1.47 & 0.49 \\
Gemini 3.1 Pro (CT) / High & 146.24 & 48.75 \\
GPT-5.3 Codex / Low & 57.79 & 19.26 \\
Doubao Seed 2.0 Lite / Default & 4.08 & 1.36 \\
GPT-OSS 120B / High & 2.19 & 0.73 \\
Doubao Seed 2.0 Lite / High & 5.98 & 1.99 \\
Doubao Seed 2.0 Lite / Low & 3.57 & 1.19 \\
Doubao Seed 2.0 Lite / Off & 4.35 & 1.45 \\
GPT-5.4 / Default & 65.16 & 21.72 \\
Qwen 3.5 Plus / Off & 6.13 & 2.04 \\
Gemini 3.1 Flash-Lite / Minimal & 5.98 & 1.99 \\
Kimi K2.5 / On & 32.88 & 10.96 \\
Kimi K2.5 / Off & 26.68 & 8.89 \\
Qwen 3.5 Flash / On & 1.64 & 0.55 \\
Qwen 3.5 35B A3B / On & 3.07 & 1.02 \\
GLM-5 / On & 46.14 & 15.38 \\
Qwen 3.5 Plus / On & 5.24 & 1.75 \\
GLM-5 / Default & 44.60 & 14.87 \\
MiniMax M2.5 / Default & 13.37 & 4.46 \\
DeepSeek V3.2 Exp / Default & 12.34 & 4.11 \\
Qwen 3.5 Flash / Off & 1.54 & 0.51 \\
Qwen 3.5 35B A3B / Off & 2.79 & 0.93 \\
GPT-OSS 120B / Low & 0.65 & 0.22 \\
GPT-5.3 Codex / Default & 25.30 & 8.43 \\
\bottomrule
\end{tabular}

%% file: experiment_assets/tables/stage_costs.tex
\begin{tabular}{@{}lrrr@{}}
\toprule
Stage & Requests & Missing token usage & Repriced tokens (\$) \\
\midrule
Forecast & 84,188 & 284 & 774.97 \\
Leakage filter & 333,009 & 879 & 114.44 \\
Diagnostic probes & 415 & 24 & 1.02 \\
Tavily search & 73,605 & n/a & -- \\
\bottomrule
\end{tabular}

%% file: experiment_assets/tables/resource_results.tex
\begin{tabular}{@{}lrrrrrr@{}}
\toprule
Configuration & Searches & Steps & Input (k) & Output (k) & Reasoning (k) & Latency (s) \\
\midrule
GPT-5.4 / High & 4.0 & 2.9 & 29.2 & 3.2 & 2.6 & 102.1 \\
Claude Opus 4.6 / Default & 3.2 & 3.1 & 35.4 & 1.6 & 0.8 & 77.6 \\
Claude Sonnet 4.6 / Default & 3.6 & 3.4 & 41.4 & 1.2 & 0.0 & 74.3 \\
GPT-OSS 120B / Default & 3.7 & 4.7 & 38.7 & 1.2 & 0.7 & 66.1 \\
Gemini 3.1 Pro (CT) / High & 3.6 & 4.3 & 46.9 & 5.7 & 4.9 & 145.7 \\
GPT-5.3 Codex / Low & 3.8 & 3.3 & 30.0 & 0.8 & 0.5 & 41.4 \\
Doubao Seed 2.0 Lite / Default & 3.4 & 4.4 & 38.7 & 1.8 & 1.5 & 88.9 \\
GPT-OSS 120B / High & 3.8 & 4.9 & 46.1 & 4.3 & 3.9 & 70.8 \\
Doubao Seed 2.0 Lite / High & 3.7 & 4.7 & 50.4 & 3.4 & 3.0 & 109.7 \\
Doubao Seed 2.0 Lite / Low & 2.9 & 4.0 & 34.7 & 1.4 & 1.1 & 70.6 \\
Doubao Seed 2.0 Lite / Off & 3.5 & 4.5 & 45.5 & 1.0 & 0.0 & 76.6 \\
GPT-5.4 / Default & 3.9 & 2.6 & 24.7 & 0.7 & 0.0 & 62.2 \\
Qwen 3.5 Plus / Off & 3.7 & 4.5 & 47.7 & 1.9 & 0.0 & 94.5 \\
Gemini 3.1 Flash-Lite / Minimal & 2.4 & 3.5 & 24.6 & 0.3 & 0.0 & 20.8 \\
Kimi K2.5 / On & 3.9 & 4.8 & 49.4 & 2.7 & 1.9 & 93.8 \\
Kimi K2.5 / Off & 3.9 & 4.9 & 43.7 & 1.5 & 0.0 & 92.3 \\
Qwen 3.5 Flash / On & 3.9 & 4.0 & 38.1 & 2.5 & 1.9 & 76.6 \\
Qwen 3.5 35B A3B / On & 3.8 & 4.1 & 40.5 & 2.4 & 1.8 & 150.7 \\
GLM-5 / On & 3.9 & 4.1 & 43.3 & 2.5 & 1.7 & 68.7 \\
Qwen 3.5 Plus / On & 4.0 & 3.5 & 37.4 & 2.2 & 1.5 & 186.2 \\
GLM-5 / Default & 4.0 & 4.1 & 41.5 & 2.5 & 1.7 & 95.0 \\
MiniMax M2.5 / Default & 3.8 & 4.9 & 42.7 & 1.7 & 1.2 & 81.7 \\
DeepSeek V3.2 Exp / Default & 3.9 & 4.9 & 44.6 & 0.9 & 0.0 & 86.7 \\
Qwen 3.5 Flash / Off & 3.0 & 3.9 & 34.5 & 2.5 & 0.0 & 51.8 \\
Qwen 3.5 35B A3B / Off & 2.9 & 3.8 & 34.0 & 2.5 & 0.0 & 91.5 \\
GPT-OSS 120B / Low & 2.2 & 3.2 & 18.0 & 0.3 & 0.1 & 37.1 \\
GPT-5.3 Codex / Default & 2.9 & 1.9 & 14.4 & 0.2 & 0.0 & 48.4 \\
\bottomrule
\end{tabular}